\documentclass[twoside,11pt]{article}

\usepackage[abbrvbib, preprint]{jmlr2e}
\usepackage{enumerate}
\usepackage{amsfonts}
\usepackage{mathrsfs}
\usepackage{amsmath,bm}
\usepackage{graphicx}
\usepackage{subfigure}
\usepackage[ruled]{algorithm2e}
\usepackage{indentfirst}
\usepackage{multirow}
\usepackage{makecell}
\usepackage{color}

\newtheorem{assumption}{Assumption}

\newcommand{\pref}[1]{{\rm(\ref{#1})}}
\makeatletter
\def\@opargbegintheorem#1#2#3{\trivlist
   \item[]{\bfseries #1\ #2\ (#3)} \itshape}
\makeatother

\jmlrheading{1}{2022}{1-48}{4/00}{10/00}{Dong20p}{Zhetong Dong, Hongwei Lin and Chi Zhou}

\ShortHeadings{Persistence B-Spline Grids: Stable Vector Representation of Persistence Diagrams }{Dong, Lin and Zhou}
\firstpageno{1}

\begin{document}

\title{Persistence B-Spline Grids: Stable Vector Representation of Persistence Diagrams Based on Data Fitting}

\author{\name Zhetong Dong\email ztdong@zju.edu.cn \\
       \AND
       \name Hongwei Lin\thanks{Corresponding author} \email hwlin@zju.edu.cn \\
	   \AND
	   \name Chi Zhou \email elonzhou@zju.edu.cn \\
       \addr School of Mathematical Sciences; \\
       State Key Lab. of CAD\& CG,\\
       Zhejiang University,\\
       Hangzhou, Zhejiang Province, P.R.China}

\editor{Kevin Murphy and Bernhard Sch{\"o}lkopf}

\maketitle

\begin{abstract}
Many attempts have been made in recent decades to integrate  machine learning (ML) and topological data analysis.
A prominent problem in applying persistent homology to ML tasks is finding a vector representation of a persistence diagram (PD),
	which is a summary diagram for representing topological features.
From the perspective of data fitting,
	a stable vector representation,
    namely,
	\emph{persistence B-spline grid} (PBSG),
	is proposed based on the efficient technique of progressive-iterative approximation for least-squares B-spline function fitting.
We theoretically prove that the PBSG method is stable with respect to the metric of 1-Wasserstein distance defined on the PD space.
The proposed method was tested on a synthetic data set, data sets of randomly generated PDs, data of a dynamical system, and 3D CAD models,
showing its effectiveness and efficiency.
\end{abstract}


    \section{Introduction}
\label{intro}
Inferring the topological structures of data at different scales is one major task in the advent of the big-data era.
\textit{Persistent homology} \citep{edelsbrunner2002topological, zomorodian2005computing} provides a robust tool to capture topological features in data.
These homological features in a dimension are summarized in a {\it persistence diagram} (PD) or a {\it barcode} \citep{ghrist2008barcodes}.
Persistent homology has also been applied successfully to deal with various practical problems in different scientific and engineering fields,
    such as biochemistry \citep{xia2015multidimensional, xia2018multiscale} and information science \citep{de2007coverage, rieck2018clique}.

The machine learning (ML) community has shown a growing interest in combining traditional ML methods and the current techniques of computational topology \citep{chazal2015subsampling, chen2016clustering, khrulkov2018geometry}.
In ML applications, classification tasks require comparing the similarity between two PDs.
Therefore,
    the distance metrics of PDs, \textit{i.e.,} the Wasserstein distance and the bottleneck distance \citep{cohen2007stability},
    play an important role.
Although some efficient algorithms \citep{kerber2017geometry} were proposed to compute the distance metrics,
    it is limited to applying ML methods and statistics on the analysis of PDs.
One of the alternatives is to extend the representation of topological features into the form of vectors that have the property of stability with respect to the distance of PDs.
A vector representation has the advantages of easy distance computation and being available for more ML tools than distance matrix form.
For instance, a vectorizing representation of PD is available to support vector machine (SVM),
    a commonly used ML tool that requires vectors as input.
This context, therefore, provides a clear motivation to represent PDs by stable vectors.

\subsection{Related Work}
Many approaches \citep{pachauri2011topology, carriere2015stable, kalivsnik2018tropical, bubenik2015statistical, reininghaus2015stable, adams2017persistence, chevyrev2020persistence} have been proposed to transform PDs into vectors.
Some attempts have been developed when studying practical applications.
For example, while studying the cortical thickness measurements of the human cortex for the study of Alzheimer's disease,
    \cite{pachauri2011topology} rasterized PDs on a regular grid to vectorize the concentration map for SVM training.
Some studies \citep{di2015comparing, adcock2016ring, kalivsnik2018tropical} represented PDs by using an algebraic polynomial or function.
For example, \cite{kalivsnik2018tropical} identified tropical coordinates on the barcode space and proved the stability of this representation with respect to the barcode metrics.
Some related studies \citep{mileyko2011probability} developed statistical measurements of the PDs,
    such as persistence landscapes \citep{bubenik2015statistical},
    the functional representation satisfying the proposed $\infty$-landscape stability and $p$-landscape stability with the same condition as the stability of the $p$-Wasserstein distance \citep{cohen2010lipschitz}.
In this paper, we will compare the performance among our methods,
    persistence landscapes~(PLs)~\citep{bubenik2015statistical}, persistence images~(PIs)~\citep{adams2017persistence}, persistence scale space kernel~(PSSK)~\citep{reininghaus2015stable}, persistence weighted Gaussian kernel~(PWGK)~\citep{kusano2018kernel}, sliced Wasserstein kernel~(SWK)~\citep{carriere2017sliced}, and persistence bag-of-words~(PBoW)~\citep{zielinski2019persistence} in some ML classification tasks.
Therefore, the related works on these competitive methods are reviewed.

\textbf{Persistence images}:
PIs proposed by \cite{adams2017persistence} combined the idea of the kernel framework in \cite{reininghaus2015stable} and the idea of ``pixel" counting in \cite{rouse2015feature}.
Specifically, the product of a Gaussian distribution and a weighted function corresponding to each PD point in the birth-persistence coordinates are added up to form a persistence surface.
By integrating the persistence surface on each pixel divided on the domain,
    a grayscale image is produced, which is often reshaped to be a vector.
In particular, the Gaussian function tends to the Dirac delta if the variance $\sigma$ approaches zero,
    and PIs degenerate into the weighted summation in \cite{rouse2015feature}.
Among their advantages, PIs allow users to assign a weight on each point in a PD, and provide an efficient and easily understandable approach to vectorize PDs for ML tasks.
Similar to the case of the kernel in \cite{reininghaus2015stable},
    PIs have 1-Wasserstein stability, and the persistence surface of a PI is additive, introducing difficulties in proving the stability results with respect to $p$-Wasserstein distance when $1<p \leq \infty$.

\textbf{Kernel methods}:
Kernel methods have been widely developed for finding an appropriate measurement of PD distance.
The main idea of the framework of kernel methods is to transform PDs into probability density functions so that the distance between the PDs can be evaluated.
Inspired by a heat diffusion problem with a Dirichlet boundary condition,
    \cite{reininghaus2015stable} proposed the PSSK.
PSSK is a framework of an additive multi-scale kernel, and has 1-Wasserstein stability.
\cite{kusano2018kernel} summarized previous studies based on kernel functions and developed the PWGK that vectorizes PDs by using kernel embedding of measures into reproducing kernel Hilbert spaces.
The stability of PWGK was proved with respect to the Hausdorff distance between data.
The SWK for PDs was proposed by \cite{carriere2017sliced}.
The points on PDs are projected on a series of lines passing through the origin, and the slopes of these lines are adjusted.
By SWK, the distance of PDs is measured as the distances between the projected points on these lines.
SWK is a negative semi-definite kernel, and it has 1-Wasserstein stability.
Furthermore, other kernel methods \citep{le2018persistence, padellini2017supervised, carriere2019metric} are being developed to be incorporated into kernel-based ML models.

\textbf{Persistence bag-of-words}:
Considering the considerable time consumption of kernel methods in practice caused by explicit computation of kernel matrix,
    \cite{zielinski2019persistence} proposed PBoW to transform a PD into a finite-dimension vector.
This method extends the conventional bag-of-words to deal with PDs.
The codebook, a finite-size representation,
    is generated by conducting clustering and obtaining the centers of clusters on the diagram into which all PDs of training samples are consolidated.
The vector representation is produced by counting the number of points closest to the codeword in the codebook in a given order.
In a mild condition, the authors gave a stable form of PBoW that has 1-Wasserstein stability.
However, the obtained vectorizing representation via PBoW depends on the training samples and the clustering methods.

\subsection{Contributions of the Present Study}
From the perspective of data fitting and approximation,
    the weighted kernel framework can be thought to interpolate with basis functions the ``importance" values.
To transform a PD into an explicit vector derived from a B-spline control grid,
    we proposed a data fitting framework based on the persistence B-spline function that approximates the ``importance" values,
    named \textit{eminence functions}.
The information of PDs is encoded into the finitely dimensional vectors produced by the control grid of the function,
    so that one can reconstruct precisely the persistence surface from the vectors.
Notably,
    while in PI~\citep{adams2017persistence},
    the persistence surface is discretized into pixels to generate the vectors.
Then, some information is ``lost" during the discretization.
The vectors produced by our method preserve the whole information of the persistence B-spline function,
    thereby improving effectiveness and efficiency.

To efficiently compute the persistence B-spline function, the technique of \textit{progressive-iterative approximation for least squares B-spline function fitting} (LSPIA) \citep{deng2014progressive} is used to robustly generate the vectors.
We also proved that the vectorizing representations have stability with respect to the $1$-Wasserstein distance under some conditions.

Generally, the proposed framework for PD representation is
(1) stable with respect to the Wasserstein distance in different choices of eminence values,
(2) efficient to compute the control grid to form the vector,
and
(3) flexible to select the ``importance" values (local eminence values and global eminence values) for different practical tasks. 
    \section{Background}
In this section,
    we provide brief introductions on B-spline function and persistent homology,
    related to the core concepts of the proposed method.

\subsection{B-Spline Function}
\label{bg:bs}
First, B-spline is briefly introduced, and readers can refer to \cite{piegl2012nurbs} for details.
A \textit{B-spline} of order $n$ is a piecewise polynomial function of degree $n-1$ with a variable $t$.
A series of non-decreasing sequence $\{\cdots \xi_{i-1}, \xi_i, \xi_{i+1},\cdots\}$ denotes \textit{knots} at which the pieces of polynomials join.
Given a sequence of knots, the B-splines of order 1 are defined by
\begin{equation}
	B_{i,1}(t) =
	\left\{
	\begin{aligned}
		& 1, \quad \xi_i \leq t < \xi_{i+1},\\
		& 0, \quad \text{otherwise}.\\
	\end{aligned}
	\right.
\end{equation}
The B-splines of $k$th-order are defined in the way of recursion by
\begin{equation}
	B_{i,k}(t) = \frac{t - \xi_i}{\xi_{i+k-1} - \xi_i} B_{i,k-1}(t) + \frac{\xi_{i+k} - t}{\xi_{i+k} - \xi_{i+k-1}}B_{i+1,k-1}(t),
\end{equation}
and thereby, $n$th-order B-spline depends on $(k-1)$-th B-spline.
The B-spline has the weight property, \textit{i.e.},
	$\sum_i B_{i,n} (t) = 1$,
for all $n\geq 1$.
Moreover, the continuity of $B_{i,n}(t)$ depends on the order $n$ and the multiplication.
Specifically, for a B-spline basis function of order $n$, if $r$ knots are coincident,
    then the first $n-r-1$ derivatives of $B_{i,n}(t)$ are continuous across that knot.
If the knots are distinct, then the first $n-2$ derivatives are continuous across each knot.
And if the distance between each knot and its predecessor,
    \textit{i.e.}, $\xi_{i+1}-\xi_i$, is the same,
    the knot vector and the corresponding B-splines are called \textit{uniform}.

A $(p,q)$-order uniform {\it B-spline function},
	of which the degree is $(p-1,q-1)$,
	is defined by
\begin{equation}
\mathcal{S}(s,t) =  \sum_{i = 1}^h \sum_{j=1}^h \mathbf{P}_{ij} B_{i,p}(s)B_{j,q}(t) \qquad s,t \in [0,1]
\end{equation}
where $\{\mathbf{P}_{ij}\}$'s are the {\it control points} in the space with the size $h\times h$,
    and the $p$th-order B-spline basis $B_{i,p}(s)$ and $q$th-order B-spline basis $B_{j,q}(t)$ are defined,
    respectively, on the knot sequences on $[0,1]$
\begin{equation*}
	S = \{\underbrace{0,\cdots,0}_{\text{p}},\xi^s_1,\cdots, \xi^s_{h-p},\underbrace{1,\cdots,1}_{\text{p}}\}
	\quad
	T = \{\underbrace{0,\cdots,0}_{\text{q}},\xi^t_1,\cdots, \xi^t_{h-q},\underbrace{1,\cdots,1}_{\text{q}}\},
\end{equation*}
which makes the property of endpoint interpolation hold. Note that, in our implementation, the bi-cubic uniform B-spline function, \textit{i.e.}, $p = q = 4$, is used for data fitting.

\subsection{Persistent Homology}
\label{bg:ph}
Persistent homology \citep{edelsbrunner2010computational, edelsbrunner2002topological} aims to infer the topological structures of the underlying shape from a point cloud sampled from it.
Persistent homology is based on formalism in algebraic topology, namely, homology~\citep{hatcher2002algebraic, kaczynski2006computational},
    which describes the ``holes'' with dimensions of a topological space in the language of algebra.
Intuitively, a homology class in the $k$-th homology group of a topological space represents a $k$-dimensional ``hole''.
To introduce the concept of PD,
    we intuitively explain the filtration based on the Vietoris-Rips (V-R) complex,
    the meaning of PDs, and the stability theorems of PD.
The textbook~\citep{edelsbrunner2010computational} provides further information.

Let $X = \{\mathbf{p}_1, \cdots, \mathbf{p}_n\}$ be a finite point set sampled from a shape in a metric space with the metric $d$.
A ball with its center $\mathbf{p}_i$ and radius $r$ is denoted as
    $B(\mathbf{p}_i,r) = \{\mathbf{x}: d(\mathbf{x}, \mathbf{p}_i)\leq r\}, i=1,\cdots, n$.
For a fixed $r\geq 0$, a $k$-simplex is defined as a subset
    $\tau = \{\mathbf{p}_{i_0}, \cdots, \mathbf{p}_{i_k}\}$ of $X$
     which satisfies that the balls of any two points intersect,
     \textit{i.e.},
     $B(\mathbf{p}_{i_{j_1}},r) \cap B(\mathbf{p}_{i_{j_2}},r) \neq \emptyset$
     for any $\mathbf{p}_{i_{j_1}}, \mathbf{p}_{i_{j_2}} \in \tau$.
The set of these simplexes with different dimensions forms a simplicial complex,
    called the V-R complex of $X$ with radius parameter $r$,
    denoted as $VR(X,r)$.
For radius parameters $r_{j_1} \leq r_{j_2}$, $VR(X,r_{j_1})$ is a subcomplex of $VR(X,r_{j_2})$,
    and a nested sequence of simplicial complexes $\mathcal{F}(X) = \{ VR(X,r_{j})\ |\ r_j \geq 0\}$ is obtained,
    called a V-R complex \textit{filtration}.

As parameter $r$ increases,
    some homological invariants emerge,
    and some disappear because of the appearance of simplexes.
Mathematically, for $r_{j_1} \leq r_{j_2}$, the inclusion between $ VR(X,r_{j_1}) $ and $ VR(X,r_{j_2}) $ induces a homomorphism $\kappa_{r_{j_1}}^{r_{j_2}}$ from the $k$-th homology group $H_k(VR(X,r_{j_1}))$ to $H_k(VR(X,r_{j_2})) $.
The $k$-th persistent homology of the filtration $\mathcal{F}(X)$ is defined by the collection of homology groups and the induced homomorphisms,
    specifically demonstrated in \cite{zomorodian2005computing}.
Intuitively, a $k$-th homological class is \textit{born} at $r_{j_1}$ if it emerges in $H_k(VR(X,r_{j_1}))$,
    and it \textit{dies} at $r_{j_2}$ if it disappears in $ H_k(VR(X,r_{j_2}))$.
Its lifetime $r_{j_2}-r_{j_1}$ is called the \textit{persistence} of the homological class.
A $k$-th PD is a multi-set of two-dimensional points $PD_k = \{(x_i, y_i)\ |\ i\in I\}$ that summarizes the birth-death pairs $(x_i,y_i) = (r_{j_1}, r_{j_2})$ of the $k$-th homological class.
On the tame condition in \cite{zomorodian2005computing}, we can assume that all PDs have finite cardinality $|I|<\infty$.

To measure the similarity of two $k$-th PDs denoted as $PD^{(1)}$ and $PD^{(2)}$,
    distance metrics are introduced. Let $\Delta =\{(x,x)\ | \ x\geq 0\}$ be the diagonal point set with infinite multiplicity,
    and $b: PD^{(1)} \cup \Delta \to PD^{(2)} \cup \Delta $ be a bijection.
The bottleneck distance between two PDs is defined by
\begin{equation}
\label{eq:bottleneck}
	W_\infty(PD^{(1)},PD^{(2)}) = \inf_{b} \sup_{P\in PD^{(1)}\cup \Delta}  \| P - b(P)\|_{\infty},
\end{equation}
and for $1\leq p \leq \infty$, the $p$-Wasserstein distance is defined by
\begin{equation}
\label{eq:wasserstein}
 W_p(PD^{(1)},PD^{(2)}) = \inf_{b}
\left(\sum_{P\in PD^{(1)}\cup \Delta} \| P - b(P)\|_{\infty}^p\right)^{1/p},
\end{equation}
where $b$ ranges over all bijections from $ PD^{(1)} \cup \Delta$ to $PD^{(2)} \cup \Delta$.
One of the most important properties of PDs is that the transformation from data to a PD is continuous with respect to the distance metrics.
As proved in \cite{cohen2007stability, chazal2014persistence},
    the bottleneck distance between two $k$-th PDs from point sets $X$ and $Y$ is controlled by the Hausdorff distance $d_H$,
    that is, $W_\infty(PD_k(X), PD_k(Y)) \leq d_H(X,Y)$.
The stability theorems with respect to $p$-Wasserstein distance are proved on some conditions of the underlying space,
    provided in \cite{cohen2010lipschitz}. 
    \section{Method and Algorithm}
\label{meandal}

\begin{figure}[!ht]
\vskip 0.1in
\begin{center}
\centerline{\includegraphics[width=11cm]{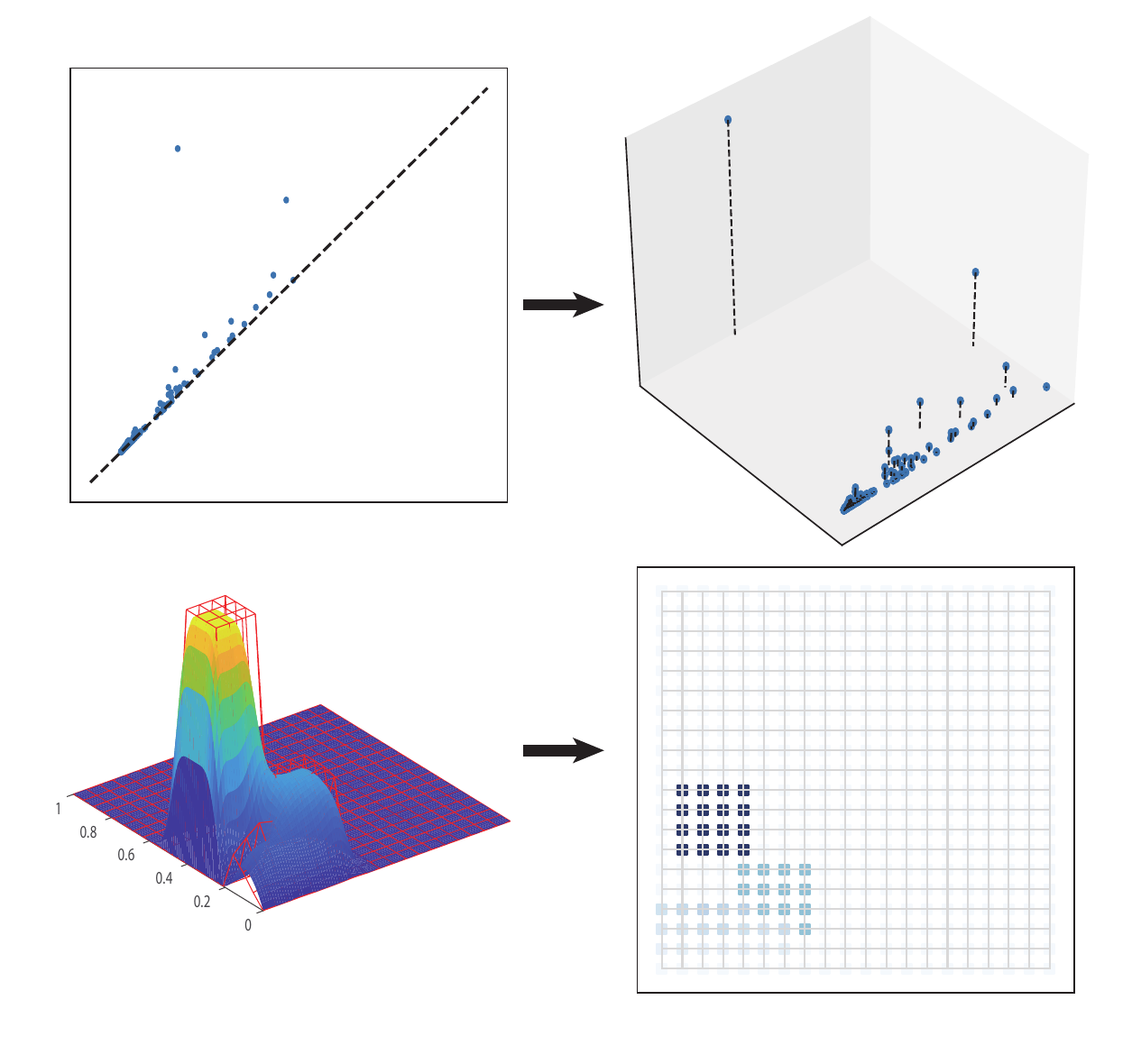}}
\caption{The pipeline of transforming a PD into a vector based on B-spline function fitting.}
\label{pip}
\end{center}
\vskip -0.1in
\end{figure}

In this section, we propose a method for transforming a PD into
    a finite-dimensional vector.
As illustrated in Fig.~\ref{pip},
    PD points are transformed from the birth-death coordinate to the birth-persistence coordinate and normalized.
Then, an \emph{eminence value} is assigned to each transformed 2D point.
Subsequently, the eminence values are fitted by a uniform bi-cubic B-spline
    function.
Finally, a finite-dimensional vector is obtained by concatenating the
    control grid of the B-spline function.

A PD with a certain dimension $k$ is a multi-set above the diagonal, \textit{i.e.},
    \begin{equation}\label{eq:pd}
        PD = \{(x_l,y_l)\in \mathbb{R}^2\ |\ y_l>x_l, l\in I\},
    \end{equation}
    where $I=\{1,2,\cdots,M\}$ is an index set.
To present our method, some assumptions should be made as follows:
 \begin{assumption} \label{assumption}
 $ $\\
    \begin{itemize}
        \item[(1)] All PDs we considered contain a finite number of points over the diagonal, \textit{i.e.}, $|I|<\infty$;
        \item[(2)] All points in all PDs are bounded,
        \footnote{Reduced 0-th PD is used to overcome the emergence of an infinitely persistent feature.}
            and denote $m$ as the maximum of the difference between upper and lower bounds of the birth coordinates and the persistence,
            \textit{i.e.}, letting $x_{min} = \min_{l\in I}(x_l)$
            and $x_{max} = \max_{l\in I} (x_l)$,
            \begin{equation*}
            \label{eq:m_def}
            m = \max\ \left\{x_{max} - x_{min}, \max_{l\in I}\ (y_l - x_l)\right\} < \infty.
            \end{equation*}
    \end{itemize}
 \end{assumption}
These assumptions are based on the theory of a tame persistent homology introduced in Section \ref{bg:ph} and practical observation.

This section is organized as follows.
In Section \ref{Sec3_1:eminence}, the coordinate transformation and eminence function are introduced.
In Section \ref{sec3_2}, the persistence B-spline grid, the key concept of our framework, is defined.
In Section \ref{sec3.3}, the algorithm based on the approximation technique LSPIA is summarized for the vectorization of PDs.

\subsection{Coordinate Transformation and Eminence Functions}
\label{Sec3_1:eminence}

As the transformation adopted in~\cite{adams2017persistence},
    PDs are transformed into the birth-persistence coordinates and normalized into a unit region.
According to Assumption~\ref{assumption}, $m$ is the maximum of the difference between upper and lower bounds of the birth coordinates and the persistence in all PDs we considered,
and the lower bound of the birth coordinates in all PDs is denoted as $c$.
\footnote{In particular, in classification tasks, $m$ and $c$ depend on the PDs of training data. For the test data, the PDs are truncated by $m$ and $c$.}
The coordinate transformation $\phi: \mathbb{R}^2 \to [0,1]^2$ is defined by
\begin{equation}
\label{eq:normalization}
\phi(x,y) = ((x-c)/m, (y-x)/m) \triangleq (s,t).
\end{equation}
The normalized birth-persistence coordinate is denoted by
$(s_l, t_l) = \phi(x_l, y_l)$ for $(x_l, y_l)\in PD$~\pref{eq:pd}.

To measure the significance of a birth-death pair in a PD,
    an \textit{eminence value} is assigned to each pair by defining different eminence functions.
The principles of designing an eminence function depend on the understanding of different persistence in a practical task.
Generally, the birth-death pair close to the diagonal is thought to be noise and marked with a small eminence value.
By contrast, the pair with large persistence is marked with a large eminence value.
Therefore, to design the eminence function, we propose two alternatives,
    namely, \textit{local eminence function} and \textit{global eminence function}.

\textbf{Local Eminence Function}:
Given a birth-death pair $(x,y)$, the local eminence function assigns to each PD point a \textit{local eminence value} that depends only on its persistence $w=y-x$.
To make the stability hold in Section \ref{stab},
    the univariate local eminence function $\mathcal{E}_L:\mathbb{R} \to \mathbb{R}$
    \begin{equation} \label{eq:local_eminence}
        \mathcal{E}_{L,PD}(x,y) \triangleq \mathcal{E}_L(y-x) = \mathcal{E}_L(w),
    \end{equation}
    should satisfy the following conditions:
\begin{enumerate}[(1)]
\item $\mathcal{E}_L(w)$ is a monotonically increasing function, \textit{i.e.}, if $w_1 < w_2$, then $\mathcal{E}_L(w_1) \leq \mathcal{E}_L(w_2)$;
\item $\mathcal{E}_L(w)$ is a bounded and differentiable function with its first-order derivative bounded,
    \textit{i.e.}, $\| \mathcal{E}_L\|_\infty < \infty$ and $\sup_{w\in \mathbb{R}} | \mathcal{E}_L^\prime (w) | < \infty$.
\end{enumerate}

\textbf{Global Eminence Function}:
To consider persistence of a PD point and the distribution of all points on a PD,
    the global eminence function assigns a \textit{global eminence value} to a PD point.
Given a PD, the global eminence function has the form
\begin{equation}
\label{eq:global_eminence}
\mathcal{E}_{G,PD}(x,y) = \sum_{l\in I} (y_l-x_l) \mathcal{K}(x-x_l,y-y_l),
\end{equation}
where $\mathcal{K}(x,y)$ is a basis function and $(x_l,y_l)\in PD$.
For stability in Section \ref{stab},
    the global eminence function is supposed to follow the principles:
\begin{enumerate}[(1)]
\item the basis function $\mathcal{K}(x,y)$ is bounded and differentiable, \textit{i.e.}, $\|\mathcal{K}(x,y)\|_\infty <\infty$;
\item the first-order partial derivatives of $\mathcal{K}(x,y)$ with respect to $x$ and $y$ are bounded,  \textit{i.e.},
$$\sup_{(x,y)\in \mathbb{R}^2} \|\nabla \mathcal{K}(x,y)\|_2 < \infty.$$
\end{enumerate}

\subsection{Persistence B-Spline Grid}
\label{sec3_2}
For each transformed coordinate $(s_l,t_l)$ in a PD,
    an eminence value $z_l$ is assigned through a local eminence function,
    \textit{i.e}.,
    \begin{equation} \label{eq:eminence_local}
        z_l = \mathcal{E}_{L,PD}(x,y) = \mathcal{E}_L(y_l - x_l),
    \end{equation}
    or through a global eminence function,
    \textit{i.e.},
    \begin{equation} \label{eq:eminence_global}
        z_l = \mathcal{E}_{G,PD}(x_l, y_l).
    \end{equation}
Then, the eminence values are fitted by a uniform bi-cubic B-spline function with $h \times h$ size of the control points as follows:
 \begin{equation} \label{eq:b_spline_function}
    \mathcal{S}(s,t) = \sum_{i = 1}^{h} \sum_{j=1}^{h}
                        \widetilde{z}_{ij} B_i(s) B_j(t),
 \end{equation}
where $h$ is the grid size satisfying $h\geq 4$, and the uniform cubic B-spline basis $B_i(s)$ and $B_j(t)$ are both defined on the uniform knot sequence
    \begin{equation}
\label{eq5}
	\xi =
    \left\{ 0,0,0,0,\frac{1}{h-3},\cdots,\frac{h-4}{h-3},1,1,1,1 \right\}.
    \end{equation}

To obtain the uniform bi-cubic B-spline function~\pref{eq:b_spline_function}
    that fits the eminence values,
    the following least-squares problem should be solved:
\begin{equation}
\label{eq:min_leastsquare}
\mathop{\min}_{ \{\widetilde{z}_{ij}\}}\ \sum_{l\in I}
    \left\|z_l - \mathcal{S}(s_l,t_l)\right\|_2^2.
\end{equation}
It is equivalent to solving the normal equation,
 \begin{equation} \label{eq:normal_equation}
    \mathbf{B^T B\widetilde{Z}} = \mathbf{B^T Z},
 \end{equation}
where
\begin{align}
	&\mathbf{\widetilde{Z}}=\left(\widetilde{z}_{11}, \widetilde{z}_{12},\cdots, \widetilde{z}_{1h},\widetilde{z}_{21}, \widetilde{z}_{22},\cdots, \widetilde{z}_{2h},\cdots,
\widetilde{z}_{h1}, \widetilde{z}_{h2},\cdots, \widetilde{z}_{hh}\right)^\mathrm{T}, \label{eq:z-vectors} \\
	&\mathbf{Z} = \left(z_1,z_2,\cdots,z_M\right)^\mathrm{T}, \label{eq:eminence_vector}
\end{align}
and the matrix $\mathbf{B}$ is the collocation matrix of the B-spline basis
    at $\{(s_l,t_l)\ |\  l\in I\}$.
By arranging the B-spline basis in a vector and denoting $B_{i}(s)B_{j}(t)$ as $B_{(i-1)\times h+j}(s,t)$, \textit{i.e.},
\begin{equation*}
\label{eq:bmatrix_row}
\begin{aligned}
     & \left[B_{1}(s)B_{1}(t), B_{1}(s)B_{2}(t),\cdots,B_{1}(s)B_{h}(t),
	   B_{2}(s)B_{1}(t), \cdots,B_{2}(s)B_{h}(t),\cdots, \right.\\
	  & \left.\qquad \qquad B_{h}(s)B_{1}(t), \cdots,B_{h}(s)B_{h}(t)\right]  \triangleq
 \left[B_1(s,t),B_2(s,t),\cdots,B_{h^2}(s,t)\right],
\end{aligned}
\end{equation*}
the collocation matrix $\mathbf{B}$ at $\{(s_l,t_l), l\in I\}$ can be written as
\begin{equation}
\label{eq:matrixB}
\mathbf{B} =
	\left[ \begin{matrix}
		B_1(s_1,t_1)& B_2(s_1,t_1) & B_3(s_1,t_1)&\cdots &B_{h^2}(s_1,t_1)\\
		B_1(s_2,t_2)& B_2(s_2,t_2) & B_3(s_2,t_2)&\cdots &B_{h^2}(s_2,t_2)\\
		\vdots& \vdots & \vdots &\cdots&\vdots\\
		\vdots& \vdots & \vdots &\cdots&\vdots\\
		B_1(s_M,t_M)& B_2(s_M,t_M) & B_3(s_M,t_M)&\cdots &B_{h^2}(s_M,t_M)
	\end{matrix}\right].
\end{equation}


We define some terminologies as follows.

\begin{definition}[Persistence B-spline function, PBSG, and vector of PBSG]
\label{def:pbsg}
$ $\\
\begin{itemize}
  \item[(1)] The uniform bi-cubic B-spline function~\pref{eq:b_spline_function} which fits the eminence values is called the \emph{persistence B-spline function}.
  \item[(2)] The control grid of the persistence B-spline function~\pref{eq:b_spline_function} is called the \emph{persistence B-spline grid (PBSG)}.
  \item[(3)] The vector $\mathbf{\widetilde{Z}}$ generated by arranging PBSG with the lexicographic order,
      shown in \pref{eq:z-vectors}, is called the \emph{vector of PBSG}.
\end{itemize}
\end{definition}

In the persistence B-spline function~\pref{eq:b_spline_function},
    the size of its control grid (PBSG) is $h \times h$.
 We call $h$ or $h \times h$ the \emph{size of the PBSG}.

\begin{definition}[Local eminence vector and global eminence vector]
\label{def:vector_of_eminence}
$ $ \\
\begin{itemize}
    \item[(1)] The vector $\mathbf{Z}$~\pref{eq:eminence_vector} is called a \emph{local eminence vector}, if its elements are taken from a local eminence function by Eq.~\pref{eq:eminence_local}.
    \item[(2)] The vector $\mathbf{Z}$~\pref{eq:eminence_vector} is called a \emph{global eminence vector}, if its elements are taken from a global eminence function by Eq.~\pref{eq:eminence_global}.
\end{itemize}
The local and global eminence vectors are both called \emph{eminence vectors}.
\end{definition}

 If a PBSG is generated by fitting a local eminence vector,
    we call it a \emph{local PBSG}.
 Similarly, if a PBSG is produced by fitting a global eminence vector,
    we call it a \emph{global PBSG}.

%
\subsection{Algorithm for PBSG}
\label{sec3.3}

As stated to obtain the solution of the least-squares fitting
    problem~\pref{eq:min_leastsquare},
    the normal equation~\pref{eq:normal_equation} should be solved.
However, the coefficient matrix of the normal equation
    \pref{eq:normal_equation} can be either non-singular or singular.
When it is singular or near singular,
    special treatments should be used to obtain a robust solution.
In this article,
    we use LSPIA~\citep{deng2014progressive,lin2018convergence} to effectively and efficiently solve the linear system~\pref{eq:normal_equation}.
Regardless of whether the coefficient matrix is singular,
    LSPIA can converge to the solution of the linear system~\pref{eq:normal_equation} in a unified format,
    without any special treatment.
When the coefficient matrix is singular,
    an infinite number of solutions are possible,
    and LSPIA converges to the solution with minimum Euclidean norm~\citep{lin2018convergence}.

LSPIA begins with an initial B-spline function
\begin{equation}
    \label{eq:s0}
    S^{(0)}(s,t) = \sum_{i = 1}^h \sum_{j = 1}^h
                \widetilde{z}^{(0)}_{ij}B_i(s)B_j(t) \quad s,t \in [0,1],
\end{equation}
where the initial control points are all zero, \textit{i.e.}, $\widetilde{z}^{(0)}_{ij}=0$.
Let $\delta^{(0)}_{l} = z_l - S^{(0)}(s_l,t_l)$ be the difference between
    the eminence values~\pref{eq:eminence_local}\pref{eq:eminence_global} and the corresponding values $S^{(0)}(s_l,t_l)$ on the function $S^{(0)}$,
    and take the initial differences for the control points as
\begin{equation}
\label{eq:delta0}
 \Delta^{(0)}_{ij} = {1\over C} \sum_{l  =1}^{M} B_i(s_l) B_j(t_l) \delta_l^{(0)},
\end{equation}
where the setting of the parameter $C$ is discussed later.
Then, adding the difference for the control points~\pref{eq:delta0} to the
    control points $\widetilde{z}^{(0)}_{ij}$,
we obtain the new control points
  $$\widetilde{z}^{(1)}_{ij} = \widetilde{z}^{(0)}_{ij} +  \Delta^{(0)}_{ij} ,$$
which then generate a new function $S^{(1)}(s,t)$.
Suppose that the $k$-th function $S^{(k)}(s_l,t_l)$ is obtained.
We calculate
\begin{equation}
\label{eq3}
\begin{aligned}
	& \delta^{(k)}_{l} = z_l - S^{(k)}(s_l,t_l),\\
	&\Delta^{(k)}_{ij} = {1\over C} \sum_{l  =1}^{M} B_i(s_l) B_j(t_l) \delta_l^{(k)},\\
	&\widetilde{z}^{(k+1)}_{ij} = \widetilde{z}^{(k)}_{ij} +  \Delta^{(k)}_{ij},
\end{aligned}
\end{equation}
and the $(k+1)$st B-spline function is generated as
\begin{equation}
\label{eq:k+1Bspline}
S^{(k+1)}(s,t) = \sum_{i = 1}^h \sum_{j = 1}^h \widetilde{z}^{(k+1)}_{ij}B_i(s)B_j(t) \quad s,t \in [0,1].
\end{equation}
In general, the iteration will stop when the given iteration time $N$
    is reached,
    or the difference error is less than a threshold $\varepsilon$,
    \textit{i.e.},
{\small
$$\left|{1\over \sqrt{M}}\left(\sqrt{\sum_{l=1}^M (\delta_l^{(k-1)})^2} - \sqrt{\sum_{l=1}^M (\delta_l^{(k)})^2}\right)\right|<\varepsilon.$$}
For the convenience to show the stability of PBSGs in Section \ref{stab},
    we set a maximum iteration time $N$ to stop the LSPIA iteration.

The parameter $C$ in \pref{eq:delta0} and \pref{eq3} is a positive constant.
If
\begin{equation}
\label{eq:C_condition}
C\geq \|\mathbf{B}^{\rm T}\|_\infty = \max_{1\leq i,j \leq h} \left\{ \sum_{l=1}^M B_i(s_l)B_j(t_l)\right\},
\end{equation}
    the LSPIA algorithm converges \citep{lin2013efficient,deng2014progressive,lin2018convergence}.
In our implementation,
    we set $C = \max_{1\leq i,j \leq h} \left\{\sum_{l = 1}^{M} B_i(s_l) B_j(t_l) \right\}$ to speed up the convergence rate of LSPIA.
Based on the result of convergence of a general LSPIA format,
    the convergence of this algorithm is proved in Appendix \ref{Appendix_B}.
Finally, we summarize the algorithm to compute the PBSG by LSPIA
    in Algorithm~\ref{alg:PBSG}.

\begin{algorithm} [!htb]
	\label{alg:PBSG}
	\SetAlgoNoLine
	\caption{Compute PBSG by LSPIA.}
	\KwIn{A PD represented by a multiset $\{(x_l, y_l)\}_{l=1}^M$,
	       the maximum iteration time $N$,
            the maximum value $m$ and the lower bound $c$ in~\pref{eq:normalization},
            and the local/global eminence function.}
    \textbf{Compute} transformed coordinates
        $(s_l, t_l) = \phi(x_l, y_l), l = 1,2\cdots, M$ by \pref{eq:normalization};\\
    \textbf{Compute} eminence values $z_l, l=1,2,\cdots,M$~\pref{eq:eminence_local}\pref{eq:eminence_global};\\
    $\widetilde{z}_{ij}^{(0)} = 0, 1\leq i,j \leq h$; \\
	\For {$k = 0$; $k \leq N$ ; $k = k + 1$}
		{
            $ S^{(k)}(s_l,t_l) = \sum_{i = 1}^h \sum_{j = 1}^h \widetilde{z}_{ij}^{(k)}B_i(s_l)B_j(t_l)$;\\
			$\delta_l^{(k)} = z_l - S^{(k)}(s_l,t_l)$, for $l = 1,2,\cdots, M$;\\	
			$\Delta_{ij}^{(k)} = {1\over C} \sum_{l  =1}^{M} B_i(s_l) B_j(t_l) \delta_l^{(k)}$;\\
			$\widetilde{z}_{ij}^{(k +1)} = \widetilde{z}_{ij}^{(k)} +\Delta_{ij}^{(k)}$;\\
		}
    \textbf{Concatenate} $\widetilde{z}_{ij}^{(N)}$ into a vector $\mathbf{\widetilde{Z}}$~\pref{eq:z-vectors} lexicographically.\\
	\KwOut{the vector of PBSG $\mathbf{\widetilde{Z}}$.}
\end{algorithm} 
    \section{Stability of Vectorizing Representation}
\label{stab}

 In this section,
    we prove the stability of PBSG with respect to 1-Wasserstein distance between PDs by the following steps:
 \begin{itemize}
   \item[(1)] We first show the stability of local and global eminence values and the transformed coordinates.
   \item[(2)] A core lemma on the LSPIA iteration method is shown to bridge the vectors of eminence values, transformed coordinates, and PBSG.
   \item[(3)] The stability of local and global PBSG is induced.
 \end{itemize}

 Let $ PD^{(1)} =\left\{\mathbf{P}_l^{(1)}=(x_l^{(1)}, y_l^{(1)})\right\}_{l=1}^M$
    be a given PD
    and $PD^{(2)} =\left\{\mathbf{P}_l^{(2)}=(x_l^{(2)}, y_l^{(2)})\right\}_{l=1}^M$ be the perturbed PD, respectively.
In the 1-Wasserstein distance between these two PDs, the mapping $b$ is the
    matching bijection that achieves the infimum, satisfying,
    $$ \mathbf{P}_l^{(2)} = b(\mathbf{P}_l^{(1)})),\ l = 1,2,\cdots,M.$$
 Therefore, we have,
 \begin{equation*}
    W_1(PD^{(1)},PD^{(2)}) =
    \sum_{l=1}^M \left\|\mathbf{P}_l^{(1)} - b(\mathbf{P}_l^{(1)})\right\|_\infty
    =
    \sum_{l=1}^M \left\|\mathbf{P}_l^{(1)} - \mathbf{P}_l^{(2)}\right\|_\infty.
 \end{equation*}

 In addition, the \emph{vectors of transformed coordinates} from
    $PD^{(i)}, i=1,2$,  given by \pref{eq:normalization},
    are denoted as,
    \begin{equation*}
         \mathbf{u}^{(i)} = \left(s_1^{(i)},\cdots,s_l^{(i)},\cdots,s_M^{(i)},t_1^{(i)},\cdots,t_l^{(i)},\cdots,t_M^{(i)}\right),
         \ i=1,2,
    \end{equation*}
 respectively.
 The \emph{eminence vectors} of
    $PD^{(i)}$,
    given by \pref{eq:eminence_vector},
    are denoted as $\mathbf{Z}^{(i)},\ i=1,2$, respectively.
 And the \emph{vectors of PBSG}~\pref{eq:z-vectors} generated from
   $PD^{(i)}$ are represented by $\mathbf{\widetilde{Z}}^{(i)},\ i=1,2$,
   respectively.
 To prove the stability of PBSG,
    we use the assumptions mentioned in Section \ref{meandal} (Assumption~\ref{assumption}).

\subsection{Stability of Eminence Values and Transformed Coordinates}
\label{sec4_1}
\textbf{Stability of local eminence values}:
In the case that the eminence values are assigned via a local eminence
    function~\pref{eq:local_eminence},
    the eminence value is determined only by the persistence of the PD point.
Thus, we have $z_l = \mathcal{E}_{L,PD}(x_l,y_l) = \mathcal{E}_L(y_l - x_l)$.
The following lemma shows that the difference between the local eminence vectors, \textit{i.e.}, $\mathbf{Z}^{(1)}$ and $\mathbf{Z}^{(2)}$,
is controlled by the multiplication of the 1-Wasserstein distance between $PD^{(i)}, i=1,2$ and a constant.

\begin{lemma}
\label{lem4.3}
Supposed that the local eminence function $\mathcal{E}_L$
 \pref{eq:local_eminence} is differentiable and has a bounded first-order derivative,
    \textit{i.e.}, $\sup_{x \in \mathbb{R}} |\mathcal{E}_L^{'}(x)| < \infty$,
    we have
\begin{equation}
\label{eq:lem4.3}
    \left\| \mathbf{Z}^{(2)} - \mathbf{Z}^{(1)} \right\|_1 \leq 2 \sup |\mathcal{E}_L^{\prime}| W_1(PD^{(1)},PD^{(2)}),
\end{equation}
where $\sup |\mathcal{E}_L^{\prime}|$ denotes $\sup_{x \in \mathbb{R}} |\mathcal{E}_L^{\prime}(x)|$.
\end{lemma}

\begin{proof}
It follows by using Lemma \ref{lem4.1} in Appendix \ref{Appendix_A} that for $1\leq l \leq M$,
\begin{equation}
\label{eq:lem4.3_1}
\begin{aligned}
\left|z_l^{(2)} - z_l^{(1)}\right| &= \left|\mathcal{E}_L(y_l^{(2)}-x_l^{(2)}) - \mathcal{E}_L(y_l^{(1)}-x_l^{(1)})\right|\\
&\leq \sup_{x \in \mathbb{R}} |\mathcal{E}_L^\prime (x)| \left| (y_l^{(2)}-x_l^{(2)})  -  (y_l^{(1)} - x_l^{(1)})\right|\\
& = \sup |\mathcal{E}_L^\prime| \left| (y_l^{(2)}-y_l^{(1)})
    -  (x_l^{(2)} - x_l^{(1)})\right|,
\end{aligned}
\end{equation}
where $\sup_{x \in \mathbb{R}} |\mathcal{E}_L^{\prime}(x)| := \sup |\mathcal{E}_L^{\prime}|$.
Because for any $a, b \in \mathbb{R}$, $|a-b|\leq |a| + |b| \leq 2 \max \{|a|, |b|\}$ holds,
    we have
\begin{equation}
\label{eq:lem4.3_2}
\begin{aligned}
\left|z_l^{(2)} - z_l^{(1)}\right| & \leq 2\sup |\mathcal{E}_L^\prime|
\max \left\{ | x_l^{(2)} - x_l^{(1)}|, | y_l^{(2)}-y_l^{(1)}|\right\}\\
& = 2\sup |\mathcal{E}_L^\prime| \left\|\mathbf{P}_l^{(1)} - \mathbf{P}_l^{(2)}\right\|_\infty,
\end{aligned}
\end{equation}
where $\mathbf{P}_l^{(1)} = \left(x_l^{(1)}, y_l^{(1)}\right)$ and $\mathbf{P}_l^{(2)} = \left(x_l^{(2)}, y_l^{(2)}\right)$.
Consider the difference of $\mathbf{Z}^{(1)}$ and $\mathbf{Z}^{(2)}$ measured by 1-norm, and it finally holds that
\begin{equation}
\label{eq:lem4.3_3}
\begin{aligned}
\left\|\mathbf{Z}^{(1)} - \mathbf{Z}^{(2)}\right\|_1 &\leq 2\sup |\mathcal{E}_L^\prime| \sum_{l=1}^M \left\|\mathbf{P}_l^{(1)} - \mathbf{P}_l^{(2)} \right\|_\infty\\
& = 2\sup |\mathcal{E}_L^\prime| W_1(PD^{(1)},PD^{(2)}).
\end{aligned}
\end{equation}
\end{proof}

\textbf{Stability of global eminence values}:
 For a global eminence function determined by all points in a PD,
    as shown in \pref{eq:global_eminence},
 the following lemma shows that the difference of the global eminence vectors under 1-norm,
 \textit{i.e.}, $\left\|\mathbf{Z}^{(1)} - \mathbf{Z}^{(2)} \right\|_1$,
 is controlled by the multiplication of the 1-Wasserstein distance between $PD^{(i)}, i=1,2$ and a constant.

\begin{lemma}
\label{lem4_3_2}
Supposed that basis function $\mathcal{K}(x,y)$ of the global eminence function in \pref{eq:global_eminence} is bounded and differentiable with bounded first-order partial derivatives,
    we have
\begin{equation}
\label{eq:lem4_3_2}
    \left\|\mathbf{Z}^{(1)} - \mathbf{Z}^{(2)} \right\|_1 \leq 2M \left(2m\sup\|\nabla\mathcal{K}\|_2 + \|\mathcal{K}\|_\infty\right) W_1(PD^{(1)}, PD^{(2)}),
\end{equation}
where {\small $\|\mathcal{K}\|_\infty = \max_{(x,y)\in \mathbb{R}^2} | \mathcal{K}(x,y)|$},
    {\small $\sup \|\nabla \mathcal{K} \|_2= \sup_{(x,y) \in \mathbb{R}^2} \| \nabla \mathcal{K}\|_2$},
    and $m$ is given in Assumption~\ref{assumption}.
\end{lemma}
The proof of Lemma \ref{lem4_3_2} is attached in Appendix \ref{Appendix_A}.

\textbf{Stability of coordinate transformation}:
Recall that birth-death coordinates are transformed to the birth-persistence coordinates through \pref{eq:normalization}.
In the following lemma,
    we show that the vector of the transformed coordinates is stable with respect to the 1-Wasserstein distance between PDs.
Specifically, the difference between the vectors of the transformed coordinates under 1-norm is controlled by the multiplication of the 1-Wasserstein distance between $PD^{(i)}, i=1,2$ and a constant.

\begin{lemma}
\label{lem4.4}
 For the vectors of transformed coordinates $\mathbf{u}^{(i)}$ from
    $PD^{(i)},\ i=1,2$, through the transformation in \pref{eq:normalization}, we have
\begin{equation}
\label{eq:lem4.4}
		\left\|\mathbf{u}^{(1)}-\mathbf{u}^{(2)}\right\|_1 \leq \frac{4}{m}W_1(PD^{(1)},PD^{(2)}),
\end{equation}
where $m$ is given in Assumption~\ref{assumption}.
\end{lemma}

\begin{proof}
Recall \pref{eq:normalization} that $s_l = (x_l - c) / m, \quad t_l = (y_l - x_l)/m$, for $1\leq l \leq M$.
Thus, we have
\begin{equation}
\begin{aligned}
\label{eq:lem4.4_1}
\left\| \mathbf{u}^{(1)} - \mathbf{u}^{(2)} \right\|_1 &= \sum_{l=1}^M \left|s_l^{(1)} - s_l^{(2)}\right| + \sum_{l=1}^M \left|t_l^{(1)} - t_l^{(2)}\right|\\
&= \frac{1}{m}\left[  \sum_{l=1}^M \left|x_l^{(1)} - x_l^{(2)}\right| + \sum_{l=1}^M \left| (y_l^{(1)} - x_l^{(1)}) - (y_l^{(2)} - x_l^{(2)})\right| \right]\\
&= \frac{1}{m}\left[  \sum_{l=1}^M \left|x_l^{(1)} - x_l^{(2)}\right| + \sum_{l=1}^M \left| (y_l^{(1)} - y_l^{(2)}) - (x_l^{(1)} - x_l^{(2)})\right| \right].
\end{aligned}
\end{equation}
It follows by the inequity $|a - b| \leq |a| + |b| \leq 2\max\{|a|, |b|\}$, for any $a, b \in \mathbb{R}$ that
\begin{equation}
\begin{aligned}
\label{eq:lem4.4_2}
\left\| \mathbf{u}^{(1)} - \mathbf{u}^{(2)}\right\|_1 &\leq \frac{1}{m}\left( \sum_{l=1}^M \left|x_l^{(1)} - x_l^{(2)}\right| + \sum_{l=1}^M \left|y_l^{(1)} - y_l^{(2)}\right| + \left|x_l^{(1)} - x_l^{(2)}\right| \right)\\
&\leq \frac{2}{m} \sum_{l=1}^M \left(\left|y_l^{(1)} - y_l^{(2)}\right| + \left|x_l^{(1)} - x_l^{(2)}\right|\right)\\
&\leq \frac{4}{m} \sum_{l=1}^M \max \left\{ \left|y_l^{(1)} - y_l^{(2)}\right|, \left|x_l^{(1)} - x_l^{(2)}\right| \right\}.
\end{aligned}
\end{equation}
Let $\mathbf{P}_l^{(1)} = \left(x_l^{(1)},y_l^{(1)}\right)$ and $\mathbf{P}_l^{(2)} = \left(x_l^{(2)},y_l^{(2)}\right)$.
It finally holds that
\begin{equation}
\label{eq:lem4.4_3}
\left\| \mathbf{u}^{(1)} - \mathbf{u}^{(2)}\right\|_1  \leq \frac{4}{m} \sum_{l=1}^M \left\|\mathbf{P}_l^{(1)} - \mathbf{P}_l^{(2)} \right\|_\infty
 = \frac{4}{m} W_1(PD^{(1)}, PD^{(2)}).
\end{equation}
\end{proof}

\subsection{Stability on LSPIA}
 In this section, we provide a lemma based on the matrix form of LSPIA
    when the iteration time $N$ is finite.
 The LSPIA iterative format shown in \pref{eq3} can be rewritten in
    a matrix form, introduced in \cite{lin2018convergence},
    as follows:
\begin{equation}
\label{iterM}
\begin{aligned}
	& \mathbf{\widetilde{Z}}_{(k+1)} = \mathbf{\widetilde{Z}}_{(k)} + \mathbf{\Lambda} \mathbf{B^\mathrm{T}}\left(\mathbf{Z} - \mathbf{B}\mathbf{\widetilde{Z}}_{(k)}\right)\\
	& \mathbf{\widetilde{Z}}_{(0)} = \mathbf{O},
\end{aligned}
\end{equation}
where $\mathbf{\widetilde{Z}}_{(k)}$ represents the vector of PBSG after $k$ iterations of LSPIA,
    $$\mathbf{\Lambda} = diag(1/C, 1/C, \cdots, 1/C)$$
 is a diagonal matrix, $C$ is an iterative parameter (a constant),
 $\mathbf{Z}$ is the eminence vector given in \pref{eq:z-vectors},
 $\mathbf{B}$ is the collocation matrix shown in \pref{eq:matrixB},
 and $\mathbf{O}$ is a zero vector.
 The iterative format in \pref{iterM} is equivalent to
\begin{equation}
\label{eq:iterformat}
\mathbf{\widetilde{Z}}_{(k+1)} = \left(\mathbf{E} - \mathbf{\Lambda B^\mathrm{T} B}\right) \mathbf{\widetilde{Z}}_{(k)} + \mathbf{\Lambda} \mathbf{B^{T}} \mathbf{Z},
\end{equation}
where $\mathbf{E}$ is an identity matrix.

In the following, we will estimate the difference between the vectors of PBSG
    $\mathbf{\widetilde{Z}}^{(i)}_{(k+1)}$ from $PD^{(i)},\ i=1,2$, under 2-norm after $k+1$ iterations of LSPIA.

\begin{lemma}
\label{lem:matrix_form}
 Let $\mathbf{B}^{(i)}$ be the collocation matrices~\pref{eq:matrixB} at the
    transformed coordinate $\mathbf{u}^{(i)},\ i=1,2$, respectively.
 For the vectors of PBSG
    $\mathbf{\widetilde{Z}}^{(i)}_{(k+1)}$ from $PD^{(i)},\ i=1,2$, after $k+1$ iterations of LSPIA,
    we have
\begin{equation}
\label{eq:lem_matrix_form}
\left\|\mathbf{\widetilde{Z}}^{(1)}_{(k+1)} - \mathbf{\widetilde{Z}}^{(2)}_{(k+1)}\right\|_2
\leq \sqrt{2}\alpha h(k+1)(2h k \frac{M}{\sqrt{C^3}}+\frac{\sqrt{M}}{C})\|\mathcal{E}\|_\infty \left\|\mathbf{u}^{(1)} - \mathbf{u}^{(2)} \right\|_2 + \frac{k+1}{\sqrt{C}} \left\|\mathbf{Z}^{(1)} - \mathbf{Z}^{(2)}\right\|_2,
\end{equation}
where $\alpha = 2(h-3)$ is the bound of $|B^\prime_i(s)|$ given in Lemma \ref{lem_derivative},
    $h$ is the size of PBSG,
    $C$ is the iterative parameter given in \pref{eq:C_condition},
    and $\mathcal{E}$ represents the local eminence function $\mathcal{E}_{L,PD}$~\pref{eq:local_eminence} or global eminence function $\mathcal{E}_{G,PD}$~\pref{eq:global_eminence}.
\end{lemma}
The proof of Lemma \ref{lem:matrix_form} is provided in Appendix \ref{Appendix_A}.

\subsection{Stability of PBSG}
\label{sec4.3}
In this section, by using the lemmas proved above,
    we provide stability results of the vector of local and global PBSG based on a finite iterations $N$ of LSPIA.

The following theorem shows that the vector of local PBSG is stable
    with respect to the 1-Wasserstein distance between PDs.

\begin{theorem}[Stability for Local Eminence Function]
\label{thm:stability}
Assume that the iterative parameter $C$ satisfies \pref{eq:C_condition} and $C\geq M$.
 For the local eminence vectors $\mathbf{Z}^{(i)}$,
    produced from a local eminence function $\mathcal{E}_L$,
    and the vectors of PBSG  $\mathbf{\widetilde{Z}}^{(i)}_{(N)}$ from $PD^{(i)},\ i=1,2$,
    we have
\begin{equation}
\label{eq:thm_stability}
\left\|\mathbf{\widetilde{Z}}^{(1)}_{(N)} - \mathbf{\widetilde{Z}}^{(2)}_{(N)}\right\|_2
\leq C_{loc} W_1(PD^{(1)},PD^{(2)}),
\end{equation}
where the \emph{stability coefficient}
\begin{equation}
\label{eq:stab_local}
C_{loc} =  \frac{2N}{\sqrt{C}} \left[ 2\sqrt{2} \frac{\alpha h }{m}(2h(N-1) + 1)\|\mathcal{E}_L\|_\infty + \sup |\mathcal{E}_L^\prime| \right],
\end{equation}
in which $\alpha = 2(h-3)$ is the bound of $|B^\prime_i(s)|$ in Lemma \ref{lem_derivative},
    $h$ is the size of PBSG, $m$ is given in Assumption~\ref{assumption},
    and $\sup|\mathcal{E}_L^\prime| = \sup_{x\in\mathbb{R}}|\mathcal{E}_L^\prime(x)|$.
\end{theorem}

\begin{proof}
Since $C\leq M$, $M/C \leq 1$.
Due to \pref{eq:lem4.2_1} in Lemma \ref{lem_norm},
we have
$$\left\|\mathbf{u}^{(1)} - \mathbf{u}^{(2)}\right\|_2 \leq \left\|\mathbf{u}^{(1)} - \mathbf{u}^{(2)}\right\|_1,\ \text{and}\ \left\|\mathbf{Z}^{(1)} - \mathbf{Z}^{(2)}\right\|_2 \leq \left\|\mathbf{Z}^{(1)} - \mathbf{Z}^{(2)}\right\|_1.$$
By substituting \pref{eq:lem4.3} and \pref{eq:lem4.4} into \pref{eq:lem_matrix_form}, \pref{eq:thm_stability} is obtained.
\end{proof}

Analogously, the following theorem shows that the vector of global PBSG is stable with respect to the 1-Wasserstein distance between PDs.

\begin{theorem}[Stability for Global Eminence Function]
\label{thm:stability_2}
Assume that the iterative parameter $C$ satisfies \pref{eq:C_condition} and $C\geq M^2$.
For the global eminence vectors $\mathbf{Z}^{(i)}$,
    produced from a global eminence function $\mathcal{E}_{G, PD} (x,y)$,
    and the vectors of PBSG  $\mathbf{\widetilde{Z}}^{(i)}_{(N)}$ from $PD^{(i)},\ i=1,2$,
    we have
\begin{equation}
\label{eq:thm_stability_2}
\left\|\mathbf{\widetilde{Z}}^{(1)}_{(N)} - \mathbf{\widetilde{Z}}^{(2)}_{(N)}\right\|_2 \leq C_{glob} W_1(PD^{(1)},PD^{(2)}),
\end{equation}
where the \emph{stability coefficient}
\begin{equation}
\label{eq:stab_global}
C_{glob} = 4\sqrt{2} \alpha h N (2h(N-1)+1)\|\mathcal{K}\|_\infty + 4Nm\sup \|\nabla \mathcal{K}\|_2 + 2N \|\mathcal{K}\|_\infty,
\end{equation}
in which $\alpha = 2(h-3)$ is the bound of $|B^\prime_i(s)|$ in Lemma \ref{lem_derivative},
    $h$ is the size of PBSG, $m$ is given in Assumption~\ref{assumption},
    {\small $\|\mathcal{K}\|_\infty = \max_{(x,y)\in \mathbb{R}^2} | \mathcal{K}(x,y)|$}, and {\small $\sup \|\nabla \mathcal{K} \|_2= \sup_{(x,y) \in \mathbb{R}^2} \| \nabla \mathcal{K}\|_2$}.
\end{theorem}

\begin{proof}
Because $C\geq M^2 \geq M$ when $M\geq 1$, $M/C \leq 1$ and $M/\sqrt{C} \leq 1$ hold.
Note that
    $\|\mathcal{E}_{G,PD}\|_\infty \leq \sum_{l=1}^M |y_l - x_l| \|\mathcal{K}(x-x_l,y-y_l)\|_\infty \leq \sum_{l=1}^M m \|\mathcal{K}\|_\infty = m M \|\mathcal{K}\|_\infty$.
Due to \pref{eq:lem4.2_1} in Lemma \ref{lem_norm},
    we have $\left\|\mathbf{u}^{(1)} - \mathbf{u}^{(2)}\right\|_2 \leq \left\|\mathbf{u}^{(1)} - \mathbf{u}^{(2)}\right\|_1$, and $\left\|\mathbf{Z}^{(1)} - \mathbf{Z}^{(2)}\right\|_2 \leq \left\|\mathbf{Z}^{(1)} - \mathbf{Z}^{(2)}\right\|_1$.
By substituting \pref{eq:lem4_3_2} and \pref{eq:lem4.4} into \pref{eq:lem_matrix_form}, \pref{eq:thm_stability_2} is obtained.
\end{proof}

Eqs.~\pref{eq:stab_local} and~\pref{eq:stab_global} explicitly give upper bounds of the stability coefficients $C_{loc}$ and $C_{glob}$.
However, it is difficult to induce the theoretical supremum for the
    stability coefficients.
In Section~\ref{sec:exp_stab}, we estimated the stability coefficients by
    computing
    \begin{equation}
    \label{eq:stab_estimate}
    \left\| \mathbf{\widetilde{Z}}^{(1)}_{(N)} -\mathbf{\widetilde{Z}}^{(2)}_{(N)} \right\|_2 /W_1(PD^{(1)},PD^{(2)}),
    \end{equation}
    using elaborately designed numerical examples.
In all the numerical examples,
    the estimation~\pref{eq:stab_estimate} of the stability coefficients are all less than $10.0$.

    \section{Experiments}
\label{exp}
 This section is organized as follows.
 In Section~\ref{sec:para_choice},
     we assessed the proposed vectorizing representation with different choices of parameters.
 In Section~\ref{sec:exp_stab}, we estimated the upper bounds of the
     stability coefficients on data sets of randomly generated PDs.
 In Section~\ref{sec:exp_classfication},
    we compare the performance of our representation and other methods, namely PIs, PLs, SWK, PWGK, PSSK, PBoW,
    in classification tasks to distinguish the dynamical systems and classify CAD models.
In Section~\ref{sec5:comp_PB_PI},
    we explained the differences between PI and our method,
    and conducted a comparison experiment to show the better performance of our method.

In all experiments,
    the 0th or 1st PDs were computed by constructing the filtration of V-R complexes of a point cloud in the space equipped with Euclidean metric by using the Python package \textit{Ripser}~\citep{bauer2017ripser}.
In classification tasks,
    five commonly used ML classifiers were used, namely,
    $k$-nearest neighbor classifier ($k$NN), random forest (RF) in \cite{breiman2001random}, gradient boosted decision trees (GBDT) in \cite{friedman2001greedy},
    logistic regression (LR),
    and linear support vector machine (LSVM) in \cite{fan2008liblinear}.
All experiments were run on a PC with an $\rm Intel^{\text{\circledR}}$ Core(TM) i7-7700 CPU@3.60 GHz$\times$8.
The data sets, MATLAB, and Python source code can be found on \url{https://github.com/ZC119/PB}.

\subsection{Effect of Parameter Choice in PBSG}
\label{sec:para_choice}
The vector of PBSG relies on multiple parameters, including the choice of eminence function and its parameter,
    the iteration time $N$ of LSPIA, and the size of PBSG $h$.
In this section, we evaluated the effect of parameter choice of PBSG on a synthetic data set containing five categories,
    and measured the classification accuracy as a quantitative index that indicates the influence of different choices of parameters.

\begin{figure}[h]
\vskip 0.1in
\begin{center}
	\subfigure{
	\begin{minipage}{2.5cm}
		\centerline{\includegraphics[width = 2.8cm]{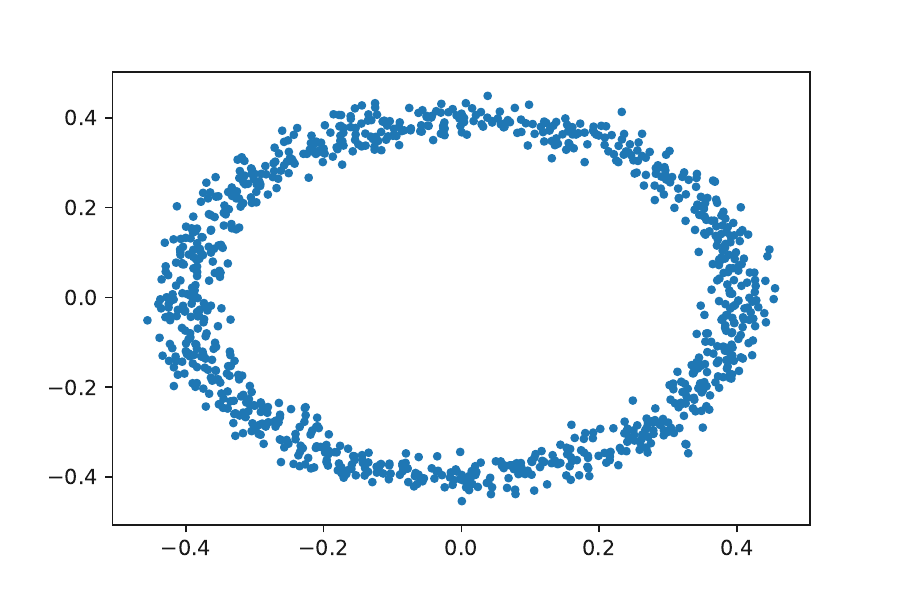}}
	\end{minipage}
	}
	\subfigure{
	\begin{minipage}{2.5cm}
		\centerline{\includegraphics[width = 2.8cm]{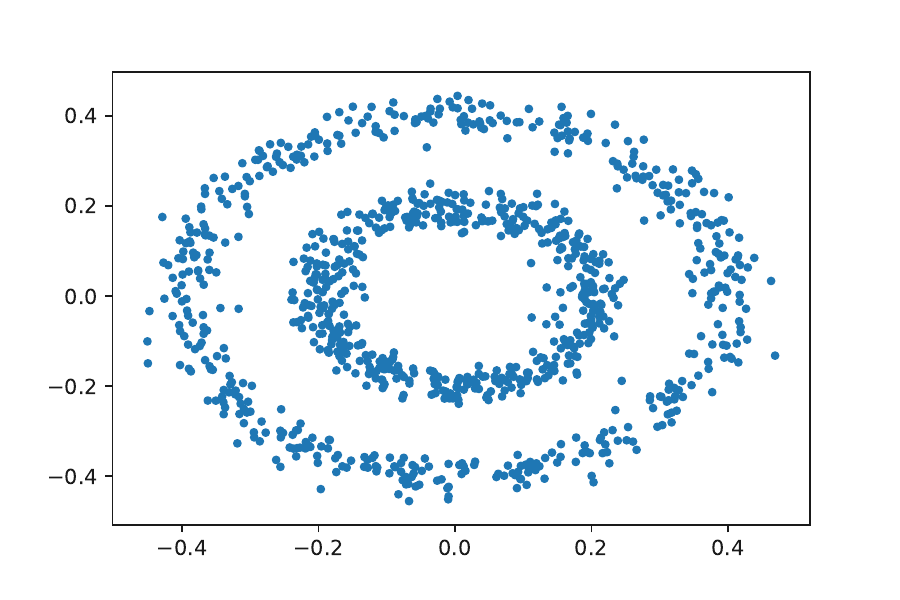}}
	\end{minipage}
	}
	\subfigure{
	\begin{minipage}{2.5cm}
		\centerline{\includegraphics[width = 2.8cm]{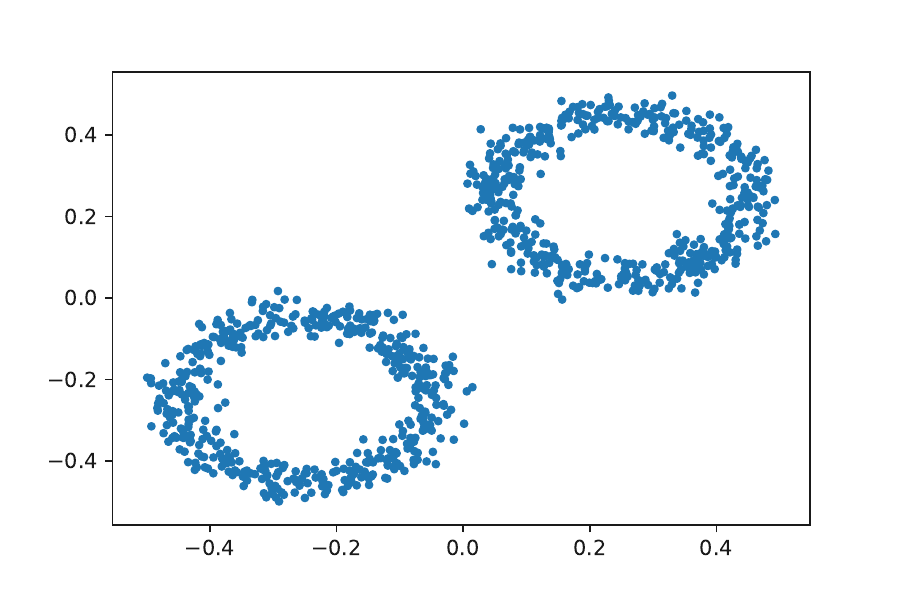}}
	\end{minipage}
	}
	\subfigure{
	\begin{minipage}{2.5cm}
		\centerline{\includegraphics[width = 2.8cm]{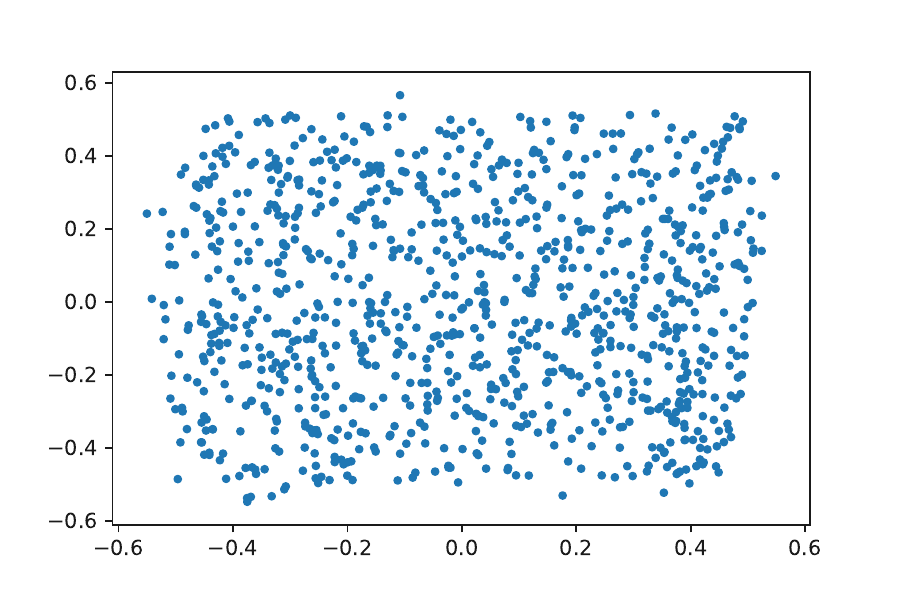}}
	\end{minipage}
	}
	\subfigure{
	\begin{minipage}{2.5cm}
		\centerline{\includegraphics[width = 2.8cm]{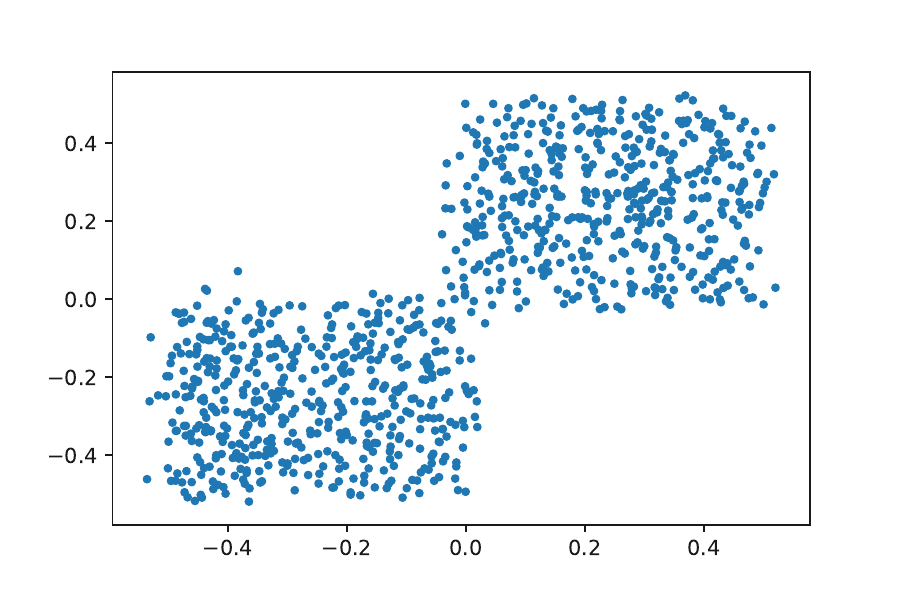}}
	\end{minipage}
	}
	\caption{The sampling points from five classes of toy data: Each sample is perturbed by a Gaussian noise with $\eta = 0.025$. }
	\label{fig:sampleimage}
\end{center}
\vskip -0.1in
\end{figure}

 As illustrated in Figure \ref{fig:sampleimage},
    the data set contains a circle with radius $0.4$;
    two concentric circles with radii $0.2$ and $0.4$, respectively;
    two disjoint circles each with radius $0.2$; a cluster of points sampled at random in the unit square;
    and two clusters of points sampled at random separately in two squares with edge length $0.5$.
 We generated $50$ images for each type,
    and sampled $1,000$ points for each image.
The sampled data were perturbed by adding Gaussian noises ($\eta = 0.025$).
 For each classifier,
    data were split into 70\% training set and 30\% test set.
 After performing 100 trials,
    the average classification accuracy was computed.
On each classifier, cross-validation was used to select the optimal parameters of classifiers.

 To measure the fitting precision of the persistence
    B-spline function~\pref{eq:b_spline_function},
    we used the root mean square (RMS) error as follows:
\begin{equation}
\label{eq:rms}
\varepsilon^{\rm RMS}= \sqrt{\frac{\sum_{l=1}^n(\delta_l^{(k)})^2}{n}},
\end{equation}
    where $n$ is the number of points in the given PD,
    and $\delta_l^{(k)}$ is presented in \pref{eq3}.
Given a collection of PDs, the average RMS error and the maximum RMS error were used to measure the fitting precision on the collection of PDs.

\textbf{Choice of local eminence function}:
 To choose a desirable local eminence function,
    we tested several commonly used candidates,
    including the linear function $\mathcal{E}_{L,PD}(x,y) = \mathcal{E}_L(y-x) = b(y-x)$,
    the \textit{arctan} function $\mathcal{E}_L(y-x) = \arctan [b(y-x)^2]$, the exponential function $\mathcal{E}_L(y-x) = e^{b(y-x)}-1$, where $b$ is a parameter,
    and the sigmoidal function
\begin{equation}
\label{eq:sigmoid}
\mathcal{E}_{L,PD}(x,y;b) = \frac{1}{1+e^{-(y-x+b)}},
\end{equation}
    where the parameter $b$ means a translation on the standard sigmoidal function.
 With the sigmoidal function,
    the eminence values of the PD points with relatively small and medium persistence varied  considerably lot,
    whereas the change of those with markedly large persistence was insignificant.
 Moreover, by using the sigmoidal function,
    the average classification accuracy of five classifiers was higher than that of other candidates.
 Therefore, the sigmoidal function was selected as a local eminence function.

\textbf{Choice of global eminence function}:
As shown in \pref{eq:global_eminence}, the global eminence function relies on a basis function and the persistence of PD points.
In our framework, we assume that a two-dimensional basis function is the multiplication of two one-dimensional basis functions,
    \textit{i.e.}, $\mathcal{K}(x,y;a_1,a_2) = \varphi(x;a_1)\cdot \varphi(y;a_2)$,
    where $a_1$ and $a_2$ are tunable parameters.
In the pretest, we set the parameters $a_1 = a_2$.
 We tested commonly employed basis functions,
    including the Gaussian basis function $\varphi(x;a) = e^{-ax^2}$, multiquadric function $\varphi(x;a) = 1+ax^2$,
    inverse multiquadric function $\varphi(x;a) = 1/\sqrt{1+ax^2}$,
    and sinc-Gaussian function  in \cite{zhang2011efficient} as follows:
\begin{equation}
\label{eq:sinc_global}
\varphi(x;a) =
\left\{
\begin{aligned}
& \frac{\sin(\pi x)}{\pi x} e^{-ax^2}, \quad x\neq 0,\\
& 1, \qquad \qquad \qquad \  x = 0,
\end{aligned}
\right.
\end{equation}
    where $a$ is a tunable parameter.
Through the pretest, the vectors of PBSG with the sinc-Gaussian function as a global function had superior performance over all other candidates on five classifiers.
In addition, \cite{zhang2011efficient} showed the $C^{\infty}$~continuity and the properties such as interpolatory property,
    partition of unity, and local support under a small tolerance.
According to these properties, the basis function and its derivative are easily verified to be bounded.
Therefore, we selected the sinc-Gaussian function~\pref{eq:sinc_global} as the basis function in the global eminence function.

\textbf{Effect of the parameters in eminence functions}:
In the pretests, we selected the sigmoidal function in~\pref{eq:sigmoid} as the local eminence function and sinc-Gaussian function in \pref{eq:sinc_global} as the basis function of the global eminence function.
We also evaluated the effect of different tunable parameters $b$ in local eminence function and $a$ in the basis function of global eminence function on the performance of classification.
Classification was conducted by setting the size of PBSG to be $20\times 20$ and the LSPIA iteration time $N=100$.
The parameter $b$ in the local eminence function was set to be $-10, -1, 1,$ and $10$.
The parameter $a$ in the global eminence function was $0.1, 1, 5,$ and $10$.
The average classification accuracy was measured on five classifiers. The average RMS error was less than $10^{-2}$ and the maximum RMS error was less than $10^{-1}$.
The experimental results are shown in Table \ref{tab:toydata}.
The classification accuracy on all classifiers was close to 100\%,
    demonstrating that the classifiers could distinguish the categories of synthetic data.
The fluctuation of accuracy on every classifier under different parameters of local and global eminence function was small (less than 1.5\%).
The classification accuracy was not sensitive to the parameter in the eminence function in the given range,
    \textit{i.e.}, $b\in [-10, 10]$, and $a\in [0.1, 10]$.

\begin{table}[ht]
\vskip 0.1in
\caption{
Classification accuracy of the vectors of PBSG with different tunable parameters in the local and global eminence functions.}
\begin{center}	
	\begin{tabular}{|c|c|c|c|c|c|}
		\hline
\textbf{Accuracy} & \textbf{$\mathbf{k}$NN} & \textbf{RF} & \textbf{GBDT} & \textbf{LR} & \textbf{LSVM} \\
		\hline
		\hline
		Local, $b=-10$  & 100\% &99.9\%& 99.1\% &98.2\% &100\% \\
		\hline
		Local, $b=-1\ $  & 100\% &99.8\%& 98.9\% &100\% &100\% \\
		\hline
		 Local, $b=1\ $  & 100\% &99.7\%& 98.8\% &100\% &100\% \\
		\hline
		Local, $b=10$ & 100\% &99.8\%& 98.8\% &100\% &100\% \\
		\hline
        \hline
        Global, $a = 0.1$ & 100\% &99.6\%& 97.8\% &99.8\% &100\% \\
        \hline
        Global, $a=1\ $ & 100\% &99.9\%& 99.4\% &100\% &100\% \\
        \hline
        Global, $a=5\ $ & 100\% &99.7\%& 98.7\% &100\% &100\% \\
        \hline
        Global, $a=10$ & 100\% &99.8\%& 98.6\% &100\% &100\% \\
        \hline
	\end{tabular}
\end{center}
\label{tab:toydata}
\vskip -0.1in
\end{table}

\begin{figure*}[!htb]
\vskip 0.1in
\begin{center}
	\subfigure[Iteration v.s. accuracy curve of local PBSG]{
		\begin{minipage}{7cm}
			\centerline{\includegraphics[width = 7.6cm]{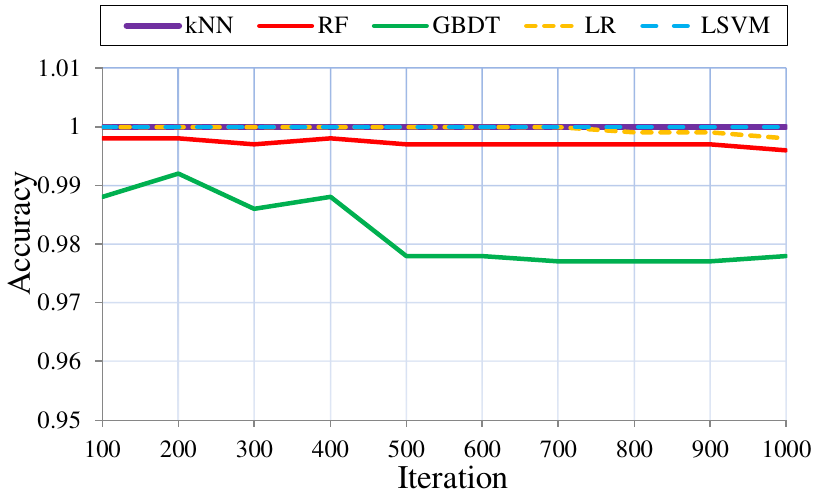}}
		\end{minipage}
    \label{fig:local_iteration}
	}
	\subfigure[PBSG size v.s. accuracy curve of local PBSG]{
		\begin{minipage}{7cm}
			\centerline{\includegraphics[width = 7cm]{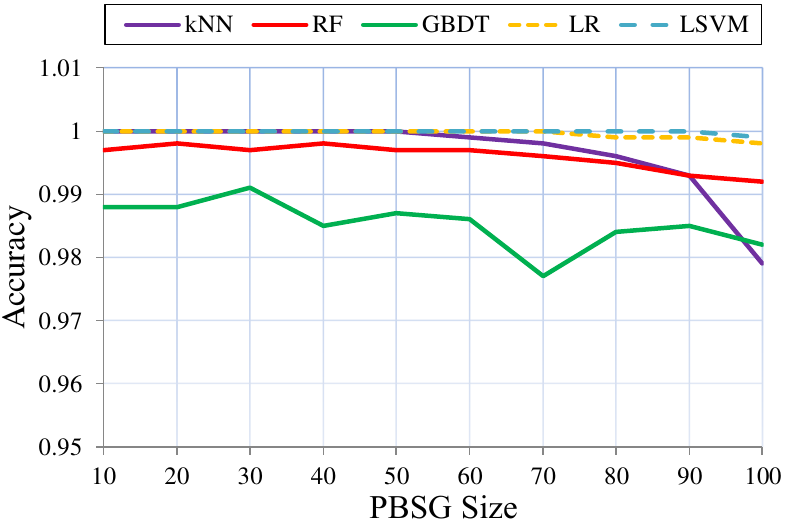}}
		\end{minipage}
    \label{fig:local_density}
	}
\subfigure[Iteration v.s. accuracy curve of global PBSG]{
		\begin{minipage}{7cm}
			\centerline{\includegraphics[width = 7cm]{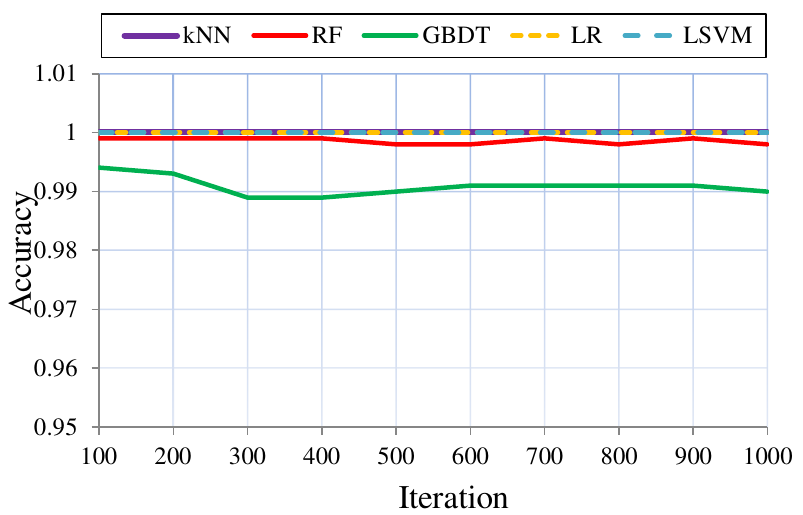}}
		\end{minipage}
    \label{fig:global_iteration}
	}
\subfigure[PBSG size v.s. accuracy curve of global PBSG]{
		\begin{minipage}{7cm}
			\centerline{\includegraphics[width = 7cm]{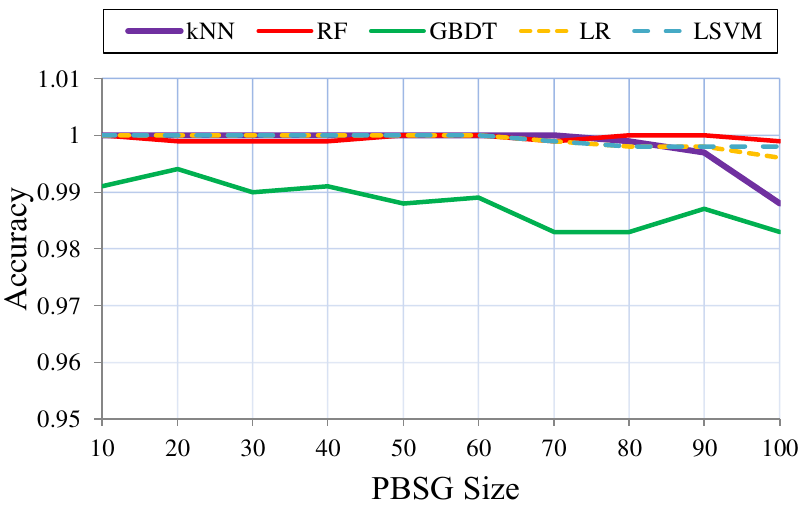}}
		\end{minipage}
    \label{fig:global_density}
	}
	\caption{
    Iteration v.s. accuracy curves and PBSG size v.s. accuracy curves of local and global PBSG on five classifiers.
}
\label{fig:curves}
\end{center}
\vskip -0.1in
\end{figure*}

\textbf{Effect of iteration time of LSPIA}:
In experiments, we set $b=10$ in the local eminence function and $a=1$ in the global eminence function.
For the iteration of LSPIA, the iteration time was increased from 100 to 1,000 at increments of 100 with the size of PBSG $20\times 20$.
The average RMS error was less than $10^{-2}$ and the maximum RMS error was less than $10^{-1}$.
For the effect of iteration time on accuracy,
    as shown in Fig.~\ref{fig:curves},
    the accuracy of the vectors of local PBSG was unchanged after 500 iterations of LSPIA on the classifier GBDT,
    and the accuracy on the other four classifiers was stable,
    as shown in Fig.~\ref{fig:local_iteration}.
 The accuracy of global PBSGs was also insensitive to the increase of iteration,
    as shown in Fig.~\ref{fig:global_iteration}.

\textbf{Effect of the size of PBSG}:
Analogously, we set $b=10$ and $a=1$.
The PBSG size increased from $10\times 10$ to $100\times 100$ at increments of 10 on each axis.
The average and maximum RMS errors were less than $10^{-2}$ and $10^{-1}$,
    respectively.
The \textit{PBSG size v.s. accuracy} curves demonstrate the effect of
    PBSG size (Fig.~\ref{fig:curves}).
The accuracy of local PBSGs fluctuated on the GBDT classifier as the PBSG size increased,
    and the accuracy on the $k$NN classifier dropped from $80\times 80$ PBSG,
    while the accuracy was stable on RF, LR, and linear SVM,
    as shown in Fig.~\ref{fig:local_density}.
Moreover, as shown in Fig.~\ref{fig:global_density},
    the accuracy of global PBSGs on the $k$NN classifier began to decrease on the $90\times 90$ PBSG,
    whereas the accuracy remained stable on the other classifiers.

In general, according to the results on the synthetic dataset,
    as LSPIA iteration and PBSG size increased,
    the classification accuracy of local and global PBSGs on the RF, LR,
    and linear SVM classifiers was more stable than that on the $k$NN and GBDT classifiers.
The accuracy did not fluctuate much on five classifiers as the iteration
    and PBSG size ascended from 100 iterations and a $10\times 10$ PBSG.
A reasonable explanation for this result is that
    the increase of iteration leads to the reduction of RMS errors
    and thus robust vectors of PBSG.
The increase in the size of PBSG causes a growth in vector dimension and a decrease in RMS error.
Consequently, we set the LSPIA iteration to be 100
    and the PBSG size to be $20\times 20$ in the subsequent experiments.

\subsection{Bound Estimation of Stability Coefficients}
\label{sec:exp_stab}
In Section \ref{stab},
    we proved the stability of the vector of PBSG with respect to 1-Wasserstein distance.
The stability coefficients~\pref{eq:stab_local}
    and \pref{eq:stab_global} mainly rely on the iterative parameter $C$, the number of iterations $N$,
    and the size of the PBSG $h$.
 In this section, the upper bounds of stability coefficients were estimated
    using numerical experiments performed on the data sets of randomly generated PDs.
 Specifically, two data sets were generated,
    each of which includes $50$ randomly generated PDs and the perturbed $50$ PDs.
 In the first collection of PDs,
    $100$ points above the diagonal were randomly generated in each PD.
 In the second collection of PDs,
    $1,000$ points in each PD were randomly generated.

\begin{figure}[!htb]
\centering
\subfigure[Average stability coefficients of local PBSG on random PDs with 100 points.]{
\begin{minipage}{7cm}
\centering
\includegraphics[width = 7cm]{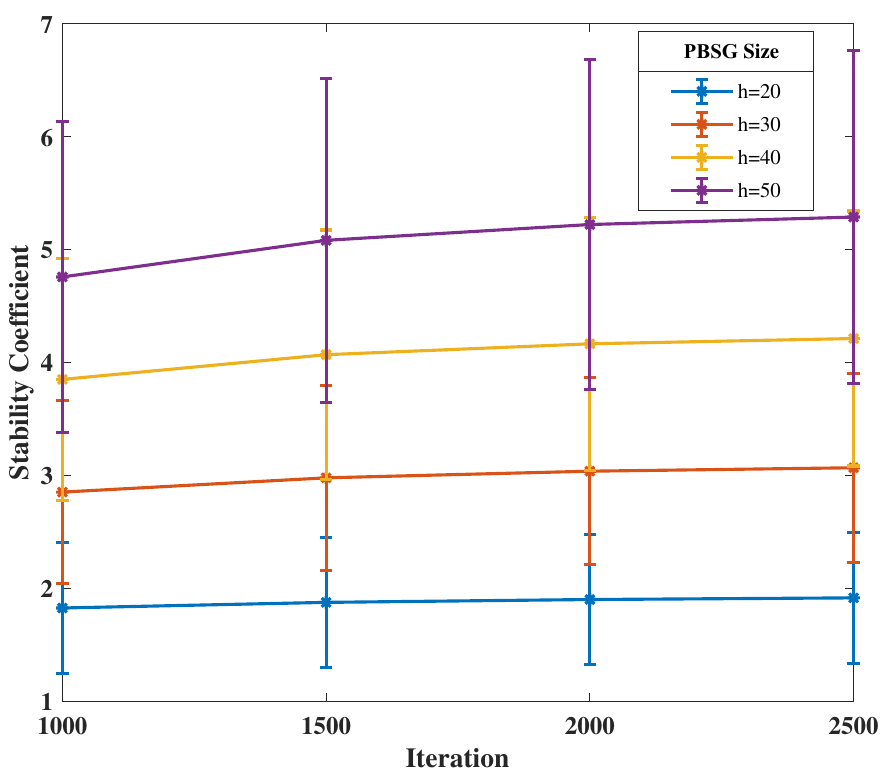}
\end{minipage}
}
\centering
\subfigure[Average stability coefficients of global PBSG on random PDs with 100 points.]{
\begin{minipage}{7cm}
\centering
\includegraphics[width = 7cm]{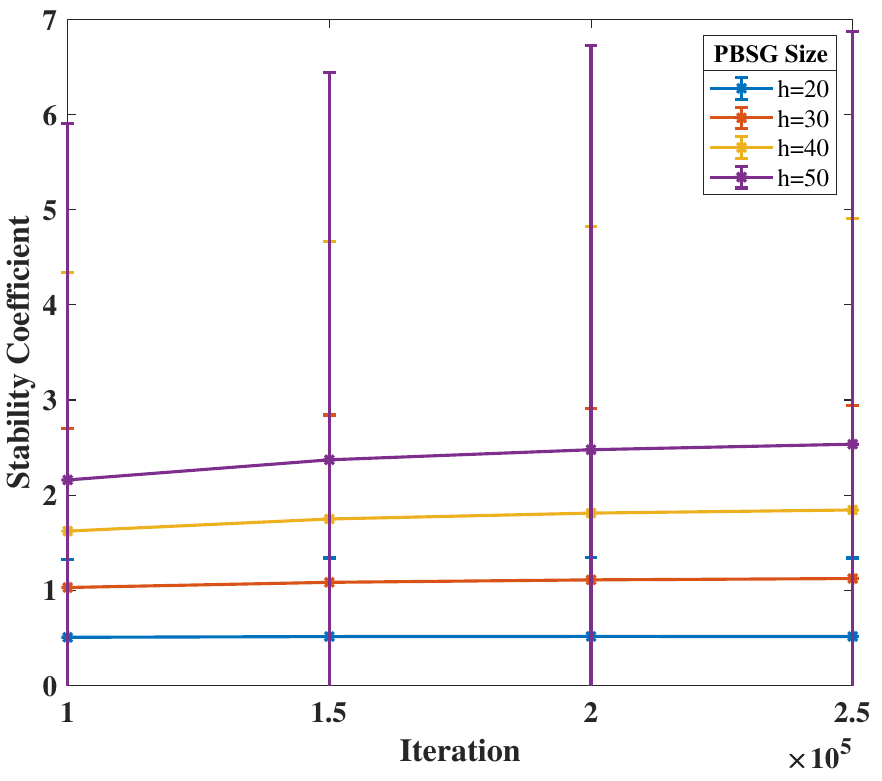}
\end{minipage}
}
\centering
\subfigure[Average stability coefficients of local PBSG on random PDs with 1000 points.]{
\begin{minipage}{7cm}
\centering
\includegraphics[width = 7cm]{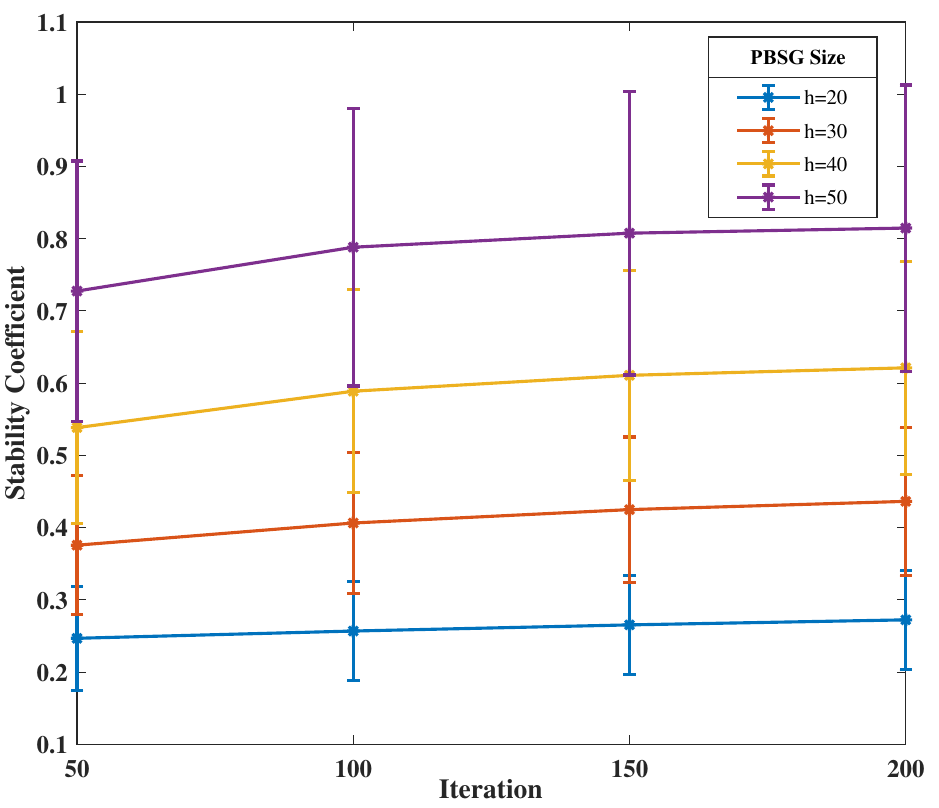}
\end{minipage}
}
\centering
\subfigure[Average stability coefficients of global PBSG on random PDs with 1000 points.]{
\begin{minipage}{7cm}
\centering
\includegraphics[width = 7cm]{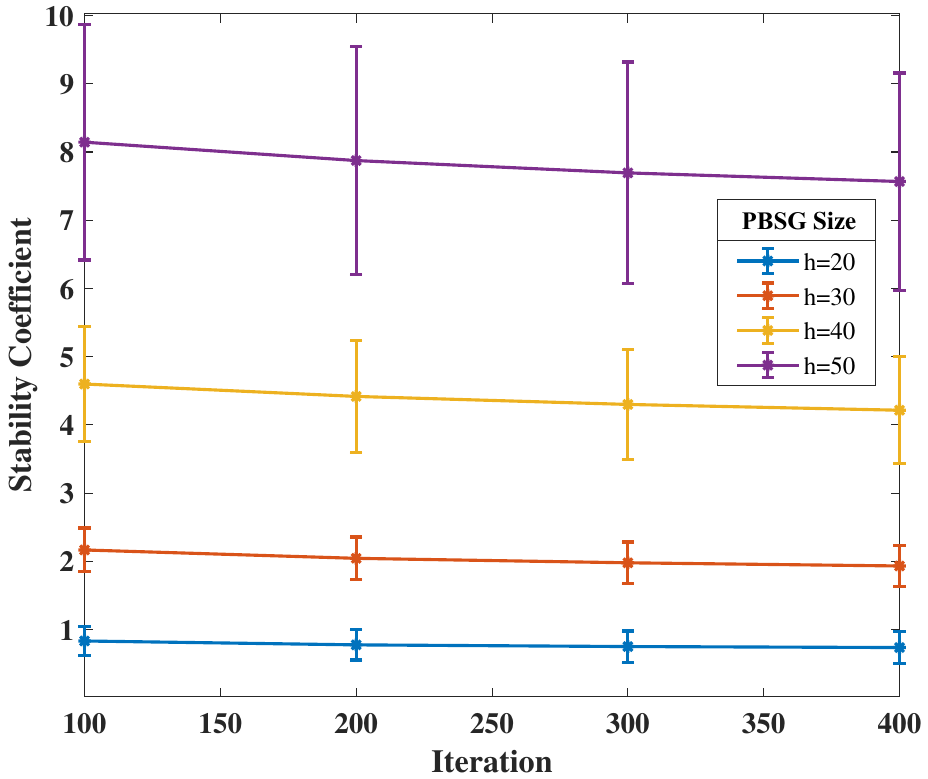}
\end{minipage}
}
\caption{Iteration v.s. stability coefficient curve with error bars under different PBSG sizes.
}
\label{fig:thm_test}
\end{figure}

Given $PD^{(1)}$ and the perturbed $PD^{(2)}$,
    let $\mathbf{\widetilde{Z}}^{(i)}_{(N)},\ i=1,2,$ be the vectors of PBSG generated after $N$ iteration of LSPIA, respectively.
In the experiments,
    we estimated the stability coefficients by computing
    $$\left\| \mathbf{\widetilde{Z}}^{(1)}_{(N)} -\mathbf{\widetilde{Z}}^{(2)}_{(N)} \right\|_2 /W_1(PD^{(1)},PD^{(2)}),$$
using the same eminence functions as those in Section \ref{sec:para_choice},
    and setting the parameter $b=1$ and $a=100$ in the local and global eminence functions, respectively.
On the first collection of PDs that contain $100$ points,
    we set $C = M$ in local PBSG
    and $C=M^2$ in global PBSG.
When $N\geq 1000$,
    the maximum RMS error of local PBSGs was not more than 0.3.
When $N \geq 10^5$, the maximum RMS error of global PBSGs was not
    more than 1.3.
On the second collection of PDs that contain $1,000$ points,
    we set $C = \max_{1\leq i,j \leq h}\left\{ \sum_{l=1}^M B_i(s_l)B_j(t_l)\right\}$ in both local and global PBSGs.
In this setting, when $N \geq 50$, the maximum RMS error of local PBSGs was less than 0.9,
    and when $N\geq 100$, the maximum RMS error of global PBSGs was less than 3.0.

As shown in Fig.~\ref{fig:thm_test},
    the averages of stability coefficient $C_{loc}$ and $C_{glob}$
    were computed with error bars.
In all of the experiments,
    the stability coefficients were all less than 10.0.
 After reaching a small RMS error, under a fixed PBSG size $h$,
    the average stability coefficients and their standard deviations were nearly unchanged with an increment of $N$.
Meanwhile, under a fixed iteration $N$,
    the average stability coefficients and their standard deviations increased with the increment of PBSG size $h$ from 20 to 50 with an increment of 10.
Specifically, in Fig.~\ref{fig:thm_test}~(a-c),
    the average stability coefficient when $h=40$ was approximately two times more than that when $h=20$.
In Fig.~\ref{fig:thm_test}~(d),
    the average stability coefficient when $h=40$ was nearly five times more than that when $h=20$.

In conclusion, the experiment demonstrates that
    after the persistence B-spline function reached a small RMS error,
    the practical stability coefficients did not sharply change as the iteration and the PBSG size increased.
Under a fixed PBSG size, the stability coefficient did not change much with
    the increment of iteration.
Under a fixed iteration time, the stability coefficient was proportional
    to the PBSG size.

\subsection{Applications on Classification}
\label{sec:exp_classfication}
To compare with other methods including computing distance matrix of
    $p$-Wasserstein distance, PIs, PLs, kernel methods (PSSK, PWGK, SWK),
    and PBoW,
    some experiments were conducted on a data set generated from a discrete dynamical system and a 3D CAD model data set.
In these experiments, software \citep{kerber2017geometry} was used
    to compute the Wasserstein distances and the bottleneck distance.
For PLs, the Persistence Landscapes Toolbox \citep{bubenik2015statistical} was used. For PIs, the open-source MATLAB codes \citep{adams2017persistence} were used.
The PIs were computed with resolutions of $20 \times 20$,
    and the parameter $\sigma$ in PIs was adjusted to guarantee the best result.

\subsubsection{Classification of Dynamical System Data with Different Parameters}
 Dynamical systems can simulate some natural phenomena.
 Different parameters in a dynamical system lead to different types of behavior.
In this experiment, PLs, PIs, PBSGs, and PDs were tested to observe their performances when classifying a complicated dynamical system with different parameters.

\begin{figure}[h]
\vskip 0.1in
\begin{center}
\centerline{\includegraphics[width = 16cm]{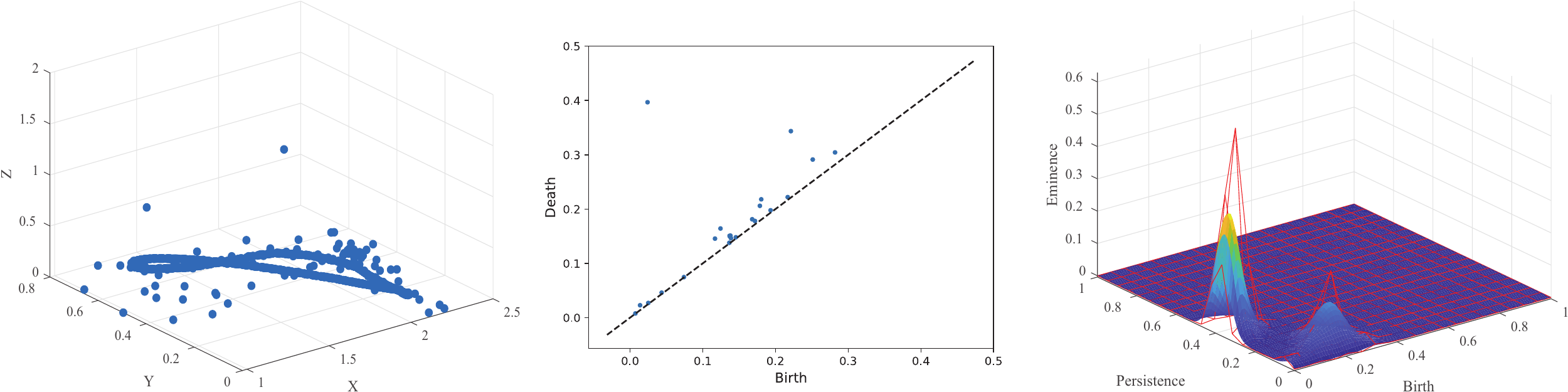}}
\caption{An example in the class of $M_0 = 3.3$. Left: the data points were generated by the 3D dynamical system with a random initial value $(X_0, Y_0, Z_0)$. Middle: the corresponding 1st PD of the data points. Right: the persistence B-spline function with the control grid marked in red.}
\label{fig:instance_3Ddynamic}
\end{center}
\vskip -0.1in
\end{figure}

The dynamical system proposed by \cite{lindstrom2002on} describes a discrete food chain model defined by
\begin{equation}
\begin{aligned}
	X_{t+1} &= \frac{M_0 X_t \exp(-Y_t)}{1+X_t \max \{ \exp(-Y_t), K(Z_t)K(Y_t)\} }\\
	Y_{t+1} &= M_1 X_t Y_t \exp (-Z_t)K(Y_t)\cdot K(M_3Y_tZ_t)\\
	Z_{t+1} &= M_2Y_t Z_t,\\
\end{aligned}
\end{equation}
where
\begin{equation*}
K(x) =
	\left\{
	\begin{aligned}
		\frac{1-\exp (-x)}{x},& \quad x \neq 0\\
		1, & \quad x = 0\\
	\end{aligned}
	\right.
\end{equation*}
The variables $X,Y$ and $Z$ are related to the trophic levels of the food chain system.
\cite{osipenko2006dynamical} studied the attractor of the system with fixed parameters $M_1 = 1.0, M_2= M_3 = 4.0$ and variable parameter $M_0 \in [3.00, 3.65]$.
Referring to Chapter 17 in \cite{osipenko2006dynamical},
    we set $M_0 = 3.0, 3.3, 3.48, 3.54,  3.57, 3.532, 3.571, 3.3701, 3.4001$,  respectively,
Thus, nine classes of results were obtained with various parameters $M_0$.
For each class, the initial values $(X_0, Y_0, Z_0)$ were generated randomly in the region $(1, 2) \times (0, 1) \times (0,1)$,
    and the system was iterated 2,000 times to obtain the points in 3D Euclidean space.
Each combination of parameters was repeated 50 times,
    and eventually 450 results were generated from the system.
In the experiment, the data were randomly split into a 70\%-30\% training-test data set,
    and the classification was conducted 10 times so that the average accuracy and its error bars were computed.
 In Fig.~\ref{fig:instance_3Ddynamic},
    we visualize an instance of a 3D point cloud of the dynamical system,
    the corresponding PD,
    the persistence B-spline function, and PBSG.

\begin{table}[h]	
\vskip 0.1in
\caption{Classification accuracy and time cost on the $k$NN classifier using different metrics ($L_1, L_2, L_\infty$): for the Wasserstein distance matrix, $L_1, L_2,$ and $L_\infty$ represent $1$-Wasserstein distance,  $2$-Wasserstein distance, and the bottleneck distance, respectively. For PLs, PIs and PBSGs, the metrics represent the Minkowski distance in vector space.}
\begin{center}
	\begin{tabular}{|c|c|c|c|}
		\hline
		{\bf Accuracy / Time } & {\bf $H_1$, $L_1$} & {\bf $H_1$, $L_2$}&{\bf $H_1$, $L_\infty$}\\
		\hline
		\hline
		$W_p$ distance matrix  & 78.8$\pm$0.6\%/667s &76.8$\pm$0.6\%/1251s & 71.9$\pm$0.7\%/983s \\
		\hline
		PLs  & 75.7$\pm$0.7\%/32s &73.4$\pm$0.6\%/48s & 71.5$\pm$0.7\%/\textbf{6s}\\
		\hline
		PIs &78.6$\pm$0.5\%/10s &80.9$\pm$0.6\%/10s &78.9$\pm$0.8\%/10s\\
		\hline
		local PBSG & 76.3$\pm$0.4\%/\textbf{7s} &76.6$\pm$0.5\%/\textbf{7s} & 65.1$\pm$0.7\% /7s \\
        \hline
        global PBSG & \textbf{83.8$\pm$0.4\%}/8s &\textbf{85.0$\pm$0.6\%}/8s & \textbf{81.3$\pm$0.4\%} /8s\\
		\hline
	\end{tabular}
\end{center}
\vskip -0.1in
\label{tab3}
\end{table}

\textbf{Comparison of $k$NN with different metrics}:
By computing $k$ nearest neighbors of a target point under a metric of the vector space,
    the $k$NN classifier determines the label of the target through voting according to those of $k$ nearest neighbors.
The metric of vector space plays an important role.
To clarify the effect on the accuracy of the $k$NN classifier with different metrics, the performance of $p$-Wasserstein ($W_p$) distance matrix, PLs, PIs, local and global PBSGs was obtained under the metrics $L_1$, $L_2$, and $L_\infty$, respectively.
For the method of computing the $p$-Wasserstein distance matrix, the chosen metrics $L_1, L_2$, and $L_\infty$ represent 1-Wasserstein, 2-Wasserstein, and the bottleneck distance, respectively.
For other methods, the metrics $L_1, L_2$, and $L_\infty$ mean the Manhattan distance, Euclidean distance, and Chebyshev distance, respectively.
As shown in Table \ref{tab3},
    the classification accuracy on the $k$NN classifier and the time cost are given regarding the $W_p$ distance matrix,
    local PBSG ($b=1$), global PBSG ($a=10$), PIs ($\sigma = 10^{-3}$),
    and PLs.
The vectors of global PBSG performed the best on the $k$NN classifier with $L_1, L_2,$ and $L_\infty$ metrics.
PIs, PLs, and PBSGs were computed efficiently,
    whereas the direct computation of the $W_p$ distance matrix required considerable time.
For PIs and PBSGs, they performed better on the $k$NN classifier by setting the metrics to be Euclidean distance.
Local PBSGs had relatively low accuracy on the $k$NN classifier with the $L_\infty$ metric.
For the global PBSGs,
    the difference of accuracy was within approximately 5.0\% among $k$NN classifiers with $L_1, L_2$, and $L_\infty$.

\textbf{Comparison among PBSGs and other competitors}:
We first measured the accuracy of local and global PBSGs under different
     parameters on five classifiers,
    shown in Table \ref{tab4_1}.
For the performance on different classifiers,
    global PBSGs ($a=1$) had the best performance on four classifiers.
 Local PBSGs also performed well on RF, GBDT, LR, and LSVM classifiers.
The accuracy did not fluctuate much by tuning the parameter of the eminence functions.

To compare local and global PBSGs with other competitive methods,
    the classification accuracy on the SVM classifier was measured, shown in Table \ref{tab4_2}.
In experiments, the hyperparameters of SVM were determined by cross-validation,
    and for PWGK, the parameters $K$ and $\rho$ were selected to be 100 and 0.01, respectively.
For the performance of different methods on SVM, global PBSGs ($a=1$) performed the best.
The performance of SWK was close to that of global PBSGs.
Local PBSGs performed better than did other methods except for SWK.

\begin{table}[ht]
\vskip 0.1in
\caption{Classification accuracy of local and global PBSGs and PIs on five classifiers.}
\begin{center}
      \begin{tabular}{|c|c|c|c|c|c|}
      \hline
      \textbf{Accuracy (\%)} &\textbf{$k$NN}&\textbf{RF}& \textbf{GBDT}&\textbf{LR}&\textbf{LSVM}\cr
      \hline
	\hline
      local PBSG ($b=1\ $) &76.7$\pm$0.4&83.5$\pm$0.5&85.2$\pm$0.6&87.4$\pm$0.5&86.9$\pm$0.5\cr\hline
      local PBSG ($b=10$) &76.2$\pm$0.5&83.5$\pm$0.4&85.6$\pm$0.6&87.7$\pm$0.6&86.5$\pm$0.7\cr\hline
      global PBSG ($a=1\ $) &83.5$\pm$0.6&\textbf{83.9$\pm$0.6}&\textbf{86.9$\pm$0.5}&\textbf{88.3$\pm$0.5}&\textbf{87.7$\pm$0.6}\cr\hline
       global PBSG ($a=10$)
       &\textbf{84.3$\pm$0.5}&83.6$\pm$0.7&85.5$\pm$0.6&87.9$\pm$0.7&87.6$\pm$0.4\cr\hline
	PIs ($\sigma = 10^{-3}$) &80.9$\pm$0.6&67.1$\pm$0.6&76.2$\pm$0.4&77.4$\pm$0.6&78.9$\pm$0.5\cr\hline
      \end{tabular}
\end{center}
\label{tab4_1}
\vskip -0.1in	
\end{table}

\begin{table}[ht]
\vskip 0.1in
\caption{Classification accuracy of local and global PBSGs, PIs, PBoW, and three kernel methods (SWK, PWGK, PSSK) on the SVM classifier.}
\begin{center}
      \begin{tabular}{|c|c|c|c|c|c|c|c|}	
      \hline
      \textbf{Acc./\%} & loc. PBSG & glob. PBSG & PIs & SWK & PWGK & PSSK &PBoW\\
	\hline
    \hline
    SVM & 86.9$\pm$0.5 & \textbf{87.7$\pm$0.6} & 78.9$\pm$0.5 & 87.2$\pm$0.3 & 80.9$\pm$0.5 & 78.2$\pm$0.6 &  82.2$\pm$0.5 \\
    \hline
      \end{tabular}
\end{center}
\label{tab4_2}
\vskip -0.1in	
\end{table}

\subsubsection{Classification of 3D CAD Models}
PDs can represent topological features with different geometric scales.
In this section, the vectorization of PDs is applied to classify some 3D CAD models.
 The benchmark data set ModelNet40 \citep{wu20153d} was used,
    which contains 12,311 CAD models classified into 40 object categories.
Initially, we computed the global PBSGs from the point clouds of all the models by uniform sampling,
    and trained the classifiers GBDT and SVM by the training data.
The accuracy of PBSG on the full benchmark was approximately 35.2\% on the GBDT classifier and approximately 28.0\% on the SVM classifier to classify shapes from 40 object categories.
Given that similar shapes may have different semantics,
    the geometrically and topologically relying methods could not recognize those shapes.
For example, the four categories, namely, bottle, bowl, cup, and flowerpot,
    have similar geometric and topological structures.
We, therefore, selected seven categories,
    namely, piano, person, airplane, radio, guitar, wardrobe, and sofa,
     with 1,970 models in total.
The 1st PDs of each category were different so that the topologically relying methods could distinguish them.

\begin{table}[ht]
	\vskip 0.1in
\caption{Classification accuracy of PIs, local and global PBSGs on five classifiers.}
\begin{center}
\begin{tabular}{|c|c|c|c|c|c|}
      \hline
      \textbf{Accuracy (\%)} &\textbf{$k$NN}&\textbf{RF}& \textbf{GBDT}&\textbf{LR}&\textbf{LSVM}\cr
      \hline
      \hline
      local PBSG ($b=10$)&83.9\%&85.4\%&85.1\%&81.0\%&82.7\%\cr\hline
      global PBSG ($a=10$)&84.9\%&85.8\%&85.6\%&82.4\%&84.8\%\cr\hline
	PIs ($\sigma = 10^{-2}$) &75.6\%&83.2\%&84.4\%&82.2\%&75.3\%\cr\hline
\end{tabular}
\end{center}
\label{tab:cl_3Dshape_1}
\vskip -0.1in	
\end{table}

\begin{table}[ht]
	\vskip 0.1in
\caption{Classification accuracy and total time cost of PBSGs and the competitive methods on the SVM classifier.
}
\begin{center}
\begin{tabular}{|c|c|c|c|c|c|}
\hline
\multicolumn{1}{|c|}{\textbf{Methods}}&\multicolumn{3}{c|}{\textbf{Accuracy on SVM}}&\multicolumn{2}{c|}{\textbf{Time Cost}}\\
	\hline
	\hline
\multicolumn{1}{|c|}{local PBSG }&\multicolumn{3}{c|}{82.7\%}&\multicolumn{2}{c|}{455s}\\
	\hline
\multicolumn{1}{|c|}{global PBSG }&\multicolumn{3}{c|}{\textbf{84.8\%}}&\multicolumn{2}{c|}{635s}\\
	\hline
\multicolumn{1}{|c|}{PIs}&\multicolumn{3}{c|}{75.3\%}&\multicolumn{2}{c|}{863s}\\
	\hline
\multicolumn{1}{|c|}{SWK}&\multicolumn{3}{c|}{83.8\%}&\multicolumn{2}{c|}{~2.5h}\\
	\hline	\multicolumn{1}{|c|}{PWGK}&\multicolumn{3}{c|}{80.9\%}&\multicolumn{2}{c|}{~2.6h}\\
	\hline \multicolumn{1}{|c|}{PSSK}&\multicolumn{3}{c|}{78.2\%}&\multicolumn{2}{c|}{~8.0h}\\
	\hline
\multicolumn{1}{|c|}{PBoW}&\multicolumn{3}{c|}{80.3\%}&\multicolumn{2}{c|}{\textbf{201s}}\\
	\hline
\end{tabular}
\end{center}
\label{tab:cl_3Dshape_2}
\vskip -0.1in	
\end{table}

As shown in Table \ref{tab:cl_3Dshape_1},
    global PBSGs achieved the best performance on five classifiers,
    and local PBSGs performed better than PIs on $k$NN, RF, GBDT, and LSVM.
In addition, as shown in Table \ref{tab:cl_3Dshape_2},
    the accuracy of global PBSGs reached the best performance in all compared methods, 1.0\% more than the second-best (SWK).
As for the time cost, PBoW took the least time (approximately 200 seconds).
Local and global PBSGs were also efficient.
In the process of PBSG computation,
     approximately 200 and 400 seconds were needed to compute the local and global eminence values, respectively,
    and approximately 200 seconds were needed in generating the PBSGs via LSPIA.
However, kernel methods (SWK, PWGK, and PSSK) needed over 2 hours in training and testing stages because the construction of kernel matrices was time-consuming.
In conclusion, the global PBSGs were efficient,
    reaching the best accuracy in the classification of 3D shapes.

\subsection{Comparison between PBSGs and PIs}
\label{sec5:comp_PB_PI}
 From the perspective of data fitting,
    PBSG and PI both fit the weighting values on the birth-death pairs in a PD with a function.
Specifically, while PI produces a quasi-interpolation surface~\citep{Cheney1995} by fitting the weights based on Gaussian kernels,
    PBSG uses a uniform bi-cubic B-spline function to fit the weights.
 However, the vectorization of the two methods is different.
On the one hand, PI discretizes the domain and integrates on each pixel
    to produce a vector,
    thereby losing some information when discretization.
 On the other hand, PBSG uses the control coefficients of the B-spline
    function to form a vector.
Given that a uniform bi-cubic B-spline function is entirely determined by
    its control coefficients,
    the vectorization of PBSG does not lose any information.
 Loosely speaking, the relationship between PBSG and PI is similar to that between
  vectorized and rasterized images.

To compare the PIs and PBSGs in practical cases,
    we designed some experiments on the data sets of the 3D dynamical system and 3D CAD models.
In these experiments,
    we set the basis function $\mathcal{K}(x,y)$ in the global eminence function of PBSG as the Gauss function:
\begin{equation}
\label{eq:Gaussian}
\mathcal{K}(x,y) = \frac{1}{2\pi \sigma^2} e^{-\frac{x^2+y^2}{2\sigma^2}},
\end{equation}
    where $\sigma$ is a parameter and $\sigma^2$ is the variance.
Thus, the weights generated at points on a PD are the same
    in PBSG and PI.
 For convenience,
    we denote by $\rm PBSG^{Gauss}$ a PBSG with Gaussian function as the basis function.
On experiments of the 3D dynamical system,
    the resolution of PIs and the size of $\rm PBSG^{Gauss}$ were both fixed as $20\times 20$,
    and the parameters $\sigma$ in PI and $\rm PBSG^{Gauss}$ were set as $10^{-1}$, $10^{-2}$, $10^{-3}$, and $10^{-4}$, respectively.
On the experiments of the 3D CAD data set,
    while the parameters $\sigma$ were fixed to be $10^{-2}$ in PI and $\rm PBSG^{Gauss}$,
    the resolution of PIs and the size of $\rm PBSG^{Gauss}$ increased from $20\times 20$ to $40\times 40$ at increments of $10$.
 In $\rm PBSG^{Gauss}$, the iteration of LSPIA was set to be 100,
    and the maximum RMS errors were less than 0.05.

\begin{table}[h]
	\vskip 0.1in
\caption{Classification accuracy of $\rm PBSG^{Gauss}$ and PI with different parameters $\sigma$ on the 3D dynamical system data set.}
\begin{center}
\begin{tabular}{|c|c|c|c|c|c|}
      \hline
      \makecell[c]{\textbf{Accuracy (\%)} \\ ($\mathbf{PBSG^{Gauss}}$/\textbf{PIs})} &\textbf{$k$NN}&\textbf{RF}& \textbf{GBDT}&\textbf{LR}&\textbf{LSVM}\cr
      \hline
      \hline      $\sigma=10^{-1}$&\textbf{81.2}/80.7&\textbf{83.6}/78.2&\textbf{85.6}/79.3&\textbf{85.6}/54.3&\textbf{84.3}/62.1\cr\hline
      $\sigma=10^{-2}$&\textbf{82.3}/80.1&\textbf{84.9}/76.5&\textbf{85.7}/79.1&\textbf{85.8}/82.3&\textbf{85.1}/83.4\cr\hline $\sigma=10^{-3}$&\textbf{87.3}/80.3&\textbf{87.1}/66.7&\textbf{89.3}/75.8&\textbf{88.8}/77.2&\textbf{88.2}/78.7\cr\hline
      $\sigma=10^{-4}$&\textbf{87.7}/73.6&\textbf{86.9}/67.9&\textbf{90.5}/75.5&\textbf{86.8}/76.5&\textbf{86.0}/76.3\cr\hline
\end{tabular}
\end{center}
\label{tab:PI_PB_dynam_sigma}
\vskip -0.1in	
\end{table}

\begin{table}[h]
	\vskip 0.1in
\caption{Classification accuracy of $\rm PBSG^{Gauss}$ and PI with different parameters $\sigma$ on the 3D CAD model data set.}
\begin{center}
\begin{tabular}{|c|c|c|c|c|c|}
      \hline
      \makecell[c]{\textbf{Accuracy (\%)} \\ ($\mathbf{PBSG^{Gauss}}$/\textbf{PIs})} &\textbf{$k$NN}&\textbf{RF}& \textbf{GBDT}&\textbf{LR}&\textbf{LSVM}\cr
      \hline
      \hline      $\sigma=10^{-1}$&\textbf{84.3}/74.0&\textbf{85.4}/74.5&\textbf{86.1}/76.2&\textbf{83.5}/70.4&\textbf{79.3}/71.6\cr\hline
      $\sigma=10^{-2}$&\textbf{83.8}/75.6&\textbf{85.2}/83.2&\textbf{86.0}/84.4&\textbf{84.5}/82.2&\textbf{76.4}/75.3\cr\hline $\sigma=10^{-3}$&\textbf{78.2}/74.8&\textbf{83.7}/81.5&\textbf{85.2}/82.0&80.9/\textbf{82.0}&69.8/\textbf{73.2}\cr\hline
      $\sigma=10^{-4}$&\textbf{77.1}/75.3&\textbf{83.8}/81.0&\textbf{85.3}/81.5&80.7/\textbf{81.9}&\textbf{63.1}/63.0\cr\hline
\end{tabular}
\end{center}
\label{tab:PI_PB_cad_sigma}
\vskip -0.1in	
\end{table}

\begin{table}[h]
	\vskip 0.1in
\caption{Classification accuracy of $\rm PBSG^{Gauss}$ and PI with different vector dimensions on the 3D dynamical system data set.}
\begin{center}
\begin{tabular}{|c|c|c|c|c|c|}
      \hline
      \makecell[c]{\textbf{Accuracy (\%)} \\ ($\mathbf{PBSG^{Gauss}}$/\textbf{PIs})} &\textbf{$k$NN}&\textbf{RF}& \textbf{GBDT}&\textbf{LR}&\textbf{LSVM}\cr
      \hline
      \hline
      $20\times 20$&\textbf{87.3}/80.3&\textbf{87.1}/66.7&\textbf{89.3}/75.8&\textbf{88.8}/77.2&\textbf{88.2}/78.7\cr\hline
      $30\times 30$&\textbf{88.2}/82.9&\textbf{85.3}/77.3&\textbf{88.7}/79.9&\textbf{90.2}/81.4&\textbf{89.6}/83.9\cr\hline $40\times 40$&\textbf{88.1}/80.5&\textbf{84.0}/77.8&\textbf{88.6}/80.4&\textbf{87.2}/80.6&\textbf{86.8}/84.2\cr\hline
\end{tabular}
\end{center}
\label{tab:PI_PB_dynam_res}
\vskip -0.1in	
\end{table}

\begin{table}[!h]
	\vskip 0.1in
\caption{Classification accuracy of $\rm PBSG^{Gauss}$ and PI with different vector dimensions on the 3D CAD model data set.}
\begin{center}
\begin{tabular}{|c|c|c|c|c|c|}
      \hline
      \makecell[c]{\textbf{Accuracy (\%)} \\ ($\mathbf{PBSG^{Gauss}}$/\textbf{PIs})} &\textbf{$k$NN}&\textbf{RF}& \textbf{GBDT}&\textbf{LR}&\textbf{LSVM}\cr
      \hline
      \hline
      $20\times 20$&\textbf{83.8}/75.6&\textbf{85.2}/83.2&\textbf{86.0}/84.4&\textbf{84.5}/82.2&\textbf{76.4}/75.3\cr\hline
      $30\times 30$&\textbf{83.9}/76.1&\textbf{84.7}/83.5&\textbf{85.3}/84.4&82.8/\textbf{83.4}&\textbf{77.7}/76.3\cr\hline $40\times 40$&\textbf{82.3}/76.0&\textbf{83.3}/83.1&\textbf{84.5}/\textbf{84.5}&81.4/\textbf{83.3}&\textbf{75.3}/\textbf{75.3}\cr\hline
\end{tabular}
\end{center}
\label{tab:PI_PB_cad_res}
\vskip -0.1in	
\end{table}

On the one hand,
    as shown in Table~\ref{tab:PI_PB_dynam_sigma},
    with the dimensions of PIs and $\rm PBSG^{Gauss}$ fixed,
    on the data set of the 3D dynamical system,
    the classification accuracy of $\rm PBSG^{Gauss}$ was higher than that of PI on all classifiers as the parameter $\sigma$ changed.
As shown in Table~\ref{tab:PI_PB_cad_sigma},
    on the 3D CAD data set,
    $\rm PBSG^{Gauss}$ performed better than did PI in most cases as the parameter $\sigma$ varied.
On the other hand,
    as listed in Table~\ref{tab:PI_PB_dynam_res},
    with the parameter $\sigma$ unchanged,
    on the data set of the 3D dynamical system,
    $\rm PBSG^{Gauss}$ had better performance than did PI on all classifiers as the PBSG size (the resolution for PIs) increased from $20\times 20$ to $40\times 40$.
In Table~\ref{tab:PI_PB_cad_res},
    the accuracy of $\rm PBSG^{Gauss}$ on the 3D CAD model data set was higher than that of PI on $k$NN, RF, GBDT,
    and LSVM as the resolution increased.
In conclusion, when the weights of PBSG were the same as that of PI,
    the performance of $\rm PBSG^{Gauss}$ was better than PI in most cases.
It should mention that the PBSG with Eq.~\pref{eq:sinc_global}
    as the basis function performs better than $\rm PBSG^{Gauss}$. 
    \section{Discussion}
\label{disc}
In this paper,
    we proposed a stable and flexible framework of PD vectorization based on a persistence B-spline function using the technique of LSPIA, \textit{i.e.}, PBSG,
    and proved the 1-Wasserstein stability of the vector of PBSG.
 The key idea of vectoring PDs is to fit the eminence values of points in
    a PD by a B-spline function,
    and reshape its control grid as a finitely dimensional vector.
 The LSPIA technique guarantees the robustness of PBSG computation.

Experiments were conducted to evaluate PBSGs and compare their performance with other competitive methods.
 As the experiments showed,
    local and global PBSGs were insensitive to the setting of LSPIA iteration and the size of PBSG.
 In practical cases, the stability coefficients did not change sharply
    as the iteration of LSPIA and the PBSG size increased.
 In the applications of classification,
    global PBSG achieved the best performance compared with other competitive approaches,
    and local PBSG also had good performance.

 As for the differences between PBSG and PI,
    the PI surface can be considered a quasi-interpolation surface based on Gaussian basis.
 The vectorization of PI relies on the discretization of the domain,
   potentially causing information loss of the PI surface.
 However, PBSG preserves the complete information of a uniform
    bi-cubic B-spline function.
 The experiments showed that PBSG performed better than did PI in most
    classification tasks.
 Moreover, the framework of PBSGs provides flexibility for users to select a
    variety of eminence functions obeying the satisfied principles,
    especially when dealing with diverse real-world applications.

    \acks{This work is supported by the National Natural Science Foundation of China under Grant nos. 61872316, 61932018,
        and the National Key R\&D Plan of China under Grant no. 2020YFB1708900.}
    \section{Appendix: Lemmas and Proofs in Section 4}
\label{Appendix_A}
The following lemmas are employed in Section \ref{stab}.
Lemma \ref{lem4.1} is an extended version of the mean value theorem in calculus.
\begin{lemma}
\label{lem4.1}
	$f: H \to \mathbb{R}$ is $C^1$ continuous on the closed interval $H\subset\mathbb{R}^n$,
		for any $\mathbf{x},\mathbf{y}\in H$,
		we have
		$$|f(\mathbf{x}) - f(\mathbf{y})| \leq \sup_{\widetilde{\mathbf{x}}\in H} || \nabla f(\widetilde{\mathbf{x}})||_2 ||\mathbf{x}-\mathbf{y}||_2 $$
	In particular,
	for a closed interval $H\subset \mathbb{R}$, if $f: H \to \mathbb{R}$,
	then
$$|f(x) - f(y)| \leq \sup_{\widetilde{x} \in H} |f^{'}(\widetilde{x})| |x-y| $$
\end{lemma}
\begin{proof}
	It follows by the mean value theorem, the Cauchy-Schwarz inequality, and the extreme value theorem. Refer to \cite{rudin1976principles}.
\end{proof}

Lemma \ref{lem4.2} gives the relationship of norms in a finitely dimensional space.

\begin{lemma}
\label{lem4.2}
	For a finite-dimensional vector space $\mathbb{R}^n$,
		let $|| \cdot||_{p_1}$ and $|| \cdot ||_{p_2}$ be given norms.
	Then there exist finite positive constants $C_m$ and $C_M$ such that
		$C_m ||\mathbf{x}||_{p_1} \leq ||\mathbf{x}||_{p_2} \leq C_M ||\mathbf{x}||_{p_1}$,
		for any $\mathbf{x} \in \mathbb{R}^n$.
Especially, the following inequities hold:
\begin{equation}
\label{eq:lem4.2_1}
\|\mathbf{x}\|_2\leq \|\mathbf{x}\|_1 \leq \sqrt{n}\|\mathbf{x}\|_2,
\end{equation}
\begin{equation}
\label{eq:lem4.2_1}
\|\mathbf{x}\|_\infty\leq \|\mathbf{x}\|_2 \leq \sqrt{n}\|\mathbf{x}\|_\infty.
\end{equation}
\end{lemma}
\begin{proof}
It follows immediately from Corollary~5.4.5 in~\cite{horn2013matrix}, and the special cases are given in Eq.~(5.4.21) in~\cite{horn2013matrix}.
\end{proof}

Lemma~\ref{lem_norm} provides some properties of the matrix norm $\|\cdot\|_2$ used in the proofs of stability theorems in Section~\ref{stab}.

\begin{lemma}
\label{lem_norm}
Let $\|\cdot\|_2$ be the matrix norm on $\mathbb{R}^{m\times n}$ induced  by the $l_2$-norm on $\mathbb{R}^n$, \
\textit{i.e.}, for $\mathbf{A} = \{a_{ij}\}_{i=1,j=1}^{m,n} \in \mathbb{R}^{m\times n}$,
$\|\mathbf{A}\|_2 = \max_{\|\mathbf{x}\|_2 = 1, \mathbf{x}\in \mathbb{R}^n} \|\mathbf{A}\mathbf{x}\|_2$.
For $\mathbf{A}, \mathbf{B} \in \mathbb{R}^{m\times n}$ and $\mathbf{x} \in \mathbb{R}^n$, the matrix norm satisfies the following properties:
\begin{enumerate}[(1)]
\item $\|c\mathbf{A}\|_2 = |c|\|\mathbf{A}\|_2$, for any $c\in \mathbb{R}$;
\item $\|\mathbf{A} + \mathbf{B}\|_2 \leq \|\mathbf{A}\|_2 + \|\mathbf{B}\|_2$;
\item $\|\mathbf{A}\mathbf{x}\|_2 \leq \|\mathbf{A}\|_2 \|\mathbf{x}\|_2$, for any $\mathbf{A}\in \mathbb{R}^{m\times n}$ and $\mathbf{x} \in \mathbb{R}^n$;
\item $\|\mathbf{A}\|_2 = \rho (\mathbf{A})$, the largest singular value of $\mathbf{A}$;
\item $\|\mathbf{A}\|_2^2 = \|\mathbf{A}^{\rm T}\mathbf{A}\|_2 = \|\mathbf{A}\mathbf{A}^{\rm T}\|_2$;
\item $\|\mathbf{A}\|_2 \leq \|\mathbf{A}\|_F$, where $\|\cdot\|_F$ is Frobenius norm, \textit{i.e.}, $\|\mathbf{A}\|_F = \sqrt{\sum_{i=1}^m\sum_{j=1}^n a_{ij}^2}$.
\end{enumerate}
\end{lemma}
\begin{proof}
It follows from the definition of matrix norm in Section~5.6 in~\cite{horn2013matrix}, Theorem~5.6.2(b), and Example~5.6.6 in~\cite{horn2013matrix}.
\end{proof}

The following lemma is provided to show that the first-order derivative of B-spline basis functions defined on the uniform knot sequence~\pref{eq5} is bounded.

\begin{lemma}
\label{lem_derivative}
The first-order derivative of cubic B-spline basis functions $B_{i,4}(s)$, $i = 1,2, \cdots h$, defined on the uniform knot sequence~\pref{eq5} is bounded, \textit{i.e.},
$|B^\prime_{i,4}(s)| \leq  \alpha = 2(h-3)$.
\end{lemma}
\begin{proof}
Due to the non-negativity and the weight property of B-spline basis functions in~\cite{piegl2012nurbs},
we have $B_{i,4}(s)\geq 0$ and $\sum_{i=1}^h B_{i,4}(s) = 1$, for $s\in [0,1]$.
It implies that $0\leq B_{i,4}(s) \leq 1$.
As given in Section~\ref{bg:bs}, because the knots are distinct and uniform on $(0,1)$, the first 2-derivatives are continuous across each knot, and therefore, $B_{i,4}(s)$ is $C^2$-continuous on $[0,1]$.
According to the formula of the first-order derivative of a $k$-degree basis function $B_{i,k+1}(s)$ that
\begin{equation*}
\label{lem_derivative_1}
B^\prime_{i,k+1}(s) = \frac{k}{\xi_{i+k} - \xi_{i}} B_{i,k}(s) - \frac{k}{\xi_{i+k+1}-\xi_{i+1}}B_{i+1,k}(s),
\end{equation*}
in our implementation where $k=3$,
the bound of the derivative is estimated on $(0,1)$ as
\begin{equation*}
\label{lem_derivative_2}
\begin{aligned}
|B^\prime_{i,4}(s)| &\leq
\left|\frac{3}{\xi_{i+3} - \xi_{i}} B_{i,3}(s)\right| + \left|\frac{3}{\xi_{i+4}-\xi_{i+1}}B_{i+1,3}(s) \right| \\
&\leq \left|\frac{3}{3/(h-3)} B_{i,3}(s)\right| +
 \left|\frac{3}{3/(h-3)}B_{i+1,3}(s)\right| \leq 2(h-3),
\end{aligned}
\end{equation*}
where $h$ is given by the uniform knot sequence~\pref{eq5}, and in our case, $h$ is the size of the control grid.
Hence, we have $|B^\prime_{i,4}(s)| \leq  \alpha = 2(h-3)$, which gives an upper bound.
\end{proof}

\begin{proof}
\textbf{of Lemma \ref{lem4_3_2}.}
First, we consider the difference between the corresponding entries in $\mathbf{Z}^{(j)}=(z_1^{(j)},z_2^{(j)},\cdots,z_M^{(j)})^T, i=1,2$, as follows:
{\small
\begin{equation}
\label{eq:lem4_3_2:1}
\left|z_i^{(1)} - z_i^{(2)}\right| = \left|\sum_{l=1}^M \left[  (y_l^{(1)} - x_l^{(1)})\mathcal{K}\left(x_i^{(1)}-x_l^{(1)}, y_i^{(1)}-y_l^{(1)}\right) - (y_l^{(2)} - x_l^{(2)})\mathcal{K}\left(x_i^{(2)}-x_l^{(2)}, y_i^{(2)}-y_l^{(2)}\right)\right] \right|.
\end{equation}
}
By adding and subtracting the same item in \pref{eq:lem4_3_2:1}, it follows by $|a+b| \leq |a| + |b|$ and $|ab| \leq |a||b|$ for any $a,b\in\mathbb{R}$ that
{\small
\begin{equation}
\label{eq:lem4_3_2:2}
\begin{aligned}
\left|z_i^{(1)} - z_i^{(2)}\right| &= \left|\sum_{l=1}^M \left[  (y_l^{(1)} - x_l^{(1)})\mathcal{K}\left(x_i^{(1)}-x_l^{(1)}, y_i^{(1)}-y_l^{(1)}\right) - (y_l^{(2)} - x_l^{(2)})\mathcal{K}\left(x_i^{(1)}-x_l^{(1)}, y_i^{(1)}-y_l^{(1)}\right) \right. \right.\\
& \left. \left. + (y_l^{(2)} - x_l^{(2)})\mathcal{K}\left(x_i^{(1)}-x_l^{(1)}, y_i^{(1)}-y_l^{(1)}\right) - (y_l^{(2)} - x_l^{(2)})\mathcal{K}\left(x_i^{(2)}-x_l^{(2)}, y_i^{(2)}-y_l^{(2)}\right)\right]\right|\\
&\leq \sum_{l=1}^M\left| (y_l^{(1)}-y_l^{(2)})-(x_l^{(1)}-y_l^{(2)})\right|
 \left|\mathcal{K}\left(x_i^{(1)}-x_l^{(1)},y_i^{(1)}-y_l^{(1)}\right)\right|\\
&+\sum_{l=1}^M \left|y_l^{(2)}-x_l^{(2)}\right|\left|  \mathcal{K}\left(x_i^{(1)}-x_l^{(1)},y_i^{(1)}-y_l^{(1)}\right)-\mathcal{K}\left(x_i^{(2)}-x_l^{(2)},y_i^{(2)}-y_l^{(2)}\right)\right|.
\end{aligned}
\end{equation}
}
Let $\mathbf{P}_l^{(j)} = (x_l^{(j)},y_l^{(j)}), j=1,2$.
In the last inequity of~\pref{eq:lem4_3_2:2}, for the first half of the inequity, because $|a-b|\leq |a|+|b|\leq 2\max\{|a|,|b|\}$ holds, we have
$$\left| (y_l^{(1)}-y_l^{(2)})-(x_l^{(1)}-y_l^{(2)})\right| \leq 2 \max\left\{|x_l^{(1)}-y_l^{(2)}|,|y_l^{(1)}-y_l^{(2)}|\right\} = 2 \|\mathbf{P}_l^{(1)}-\mathbf{P}_l^{(2)}\|_\infty.$$
Meanwhile, it follows that
$$\left|\mathcal{K}(x_i^{(1)}-x_l^{(1)},y_i^{(1)}-y_l^{(1)})\right| \leq \max_{(x,y)\in \mathbb{R}^2} | \mathcal{K}(x,y)| = \|\mathcal{K}\|_\infty.$$
For the second half of the inequity in~\pref{eq:lem4_3_2:2},
it follows by using the fundamental theorem of calculus given in Lemma~\ref{lem4.1} in Appendix~\ref{Appendix_A} that
\begin{equation}
\label{eq:lem4_3_2:3}
\left|\mathcal{K}(\mathbf{x}^{(1)})-\mathcal{K}(\mathbf{x}^{(2)})\right| \leq \sup_{(x,y) \in \mathbb{R}^2} \| \nabla \mathcal{K}(x,y)\|_2 \|\mathbf{x}^{(1)} - \mathbf{x}^{(2)}\|_2,
\end{equation}
where $\mathbf{x}^{(1)} = (x_i^{(1)}-x_l^{(1)},y_i^{(1)}-y_l^{(1)})$ and $\mathbf{x}^{(2)} = (x_i^{(2)}-x_l^{(2)},y_i^{(2)}-y_l^{(2)})$.
For convenience, let {\small $\sup \|\nabla \mathcal{K} \|_2= \sup_{(x,y) \in \mathbb{R}^2} \| \nabla \mathcal{K}\|_2$}.
Since $\|\mathbf{x}\|_2\leq \|\mathbf{x}\|_1$ for any $\mathbf{x}\in \mathbb{R}^2$, it induces from \pref{eq:lem4_3_2:3} that
\begin{equation}
\label{eq:lem4_3_2:4}
\left|\mathcal{K}(\mathbf{x}^{(1)})-\mathcal{K}(\mathbf{x}^{(2)})\right| \leq \sup \|\nabla \mathcal{K} \|_2 \|\mathbf{x}^{(1)} - \mathbf{x}^{(2)}\|_1.
\end{equation}
In addition, $\left|y_l^{(2)}-x_l^{(2)}\right|$ is bounded by $m$ (Assumption~\ref{assumption}).
It is induced from~\pref{eq:lem4_3_2:2} that
\begin{equation}
\label{eq:lem4_3_2:5}
\left|z_i^{(1)} - z_i^{(2)}\right| \leq 2\|\mathcal{K}\|_\infty\sum_{l=1}^M\|\mathbf{P}_l^{(1)} - \mathbf{P}_l^{(2)}\|_\infty + m\sup \|\nabla \mathcal{K} \|_2 \sum_{l=1}^M \|\mathbf{x}^{(1)} - \mathbf{x}^{(2)}\|_1.
\end{equation}
Notice that
\begin{equation*}
    \begin{aligned}
       \|\mathbf{x}^{(1)} - \mathbf{x}^{(2)}\|_1 & = \left|(x_i^{(1)}-x_l^{(1)})- (x_i^{(2)}-x_l^{(2)})\right| + \left|(y_i^{(1)}-y_l^{(1)}) - (y_i^{(2)}-y_l^{(2)})\right| \\
       & = \left|(x_i^{(1)}-x_i^{(2)})- (x_l^{(1)}-x_l^{(2)})\right| + \left|(y_i^{(1)}-y_i^{(2)}) - (y_l^{(1)}-y_l^{(2)})\right|.
    \end{aligned}
\end{equation*}
Hence, according to the inequity
    $|a-b| \leq |a|+|b|\leq 2\max\{|a|,|b|\}$ for any $a,b\in\mathbb{R}$,
    we have
\begin{equation*}
    \begin{aligned}
       \|\mathbf{x}^{(1)} - \mathbf{x}^{(2)}\|_1 & \leq |x_i^{(1)}-x_i^{(2)}| + |x_l^{(1)}-x_l^{(2)}| + |y_i^{(1)}-y_i^{(2)}| + |y_l^{(1)}-y_l^{(2)}| \\
       & \leq 2(\max \left\{|x_i^{(1)}-x_i^{(2)}|,|y_i^{(1)}-y_i^{(2)}|\right\}+\max \left\{|x_l^{(1)}-x_l^{(2)}|,
       |y_l^{(1)}-y_l^{(2)}|\right\} \\
       & = 2\left(\|\mathbf{P}_i^{(1)} - \mathbf{P}_i^{(2)}\|_\infty + \|\mathbf{P}_l^{(1)} - \mathbf{P}_l^{(2)}\|_\infty\right).
    \end{aligned}
\end{equation*}
According to the matching determined by the bijection $b$,
$$W_1(PD^{(1)},PD^{(2)}) = \sum_{l=1}^M\|\mathbf{P}_l^{(1)} - \mathbf{P}_l^{(2)}\|_\infty.$$
It follows by \pref{eq:lem4_3_2:5} that
{\small
\begin{equation}
\label{eq:lem4_3_2:6}
\begin{aligned}
\left|z_i^{(1)} - z_i^{(2)}\right| & \leq 2\|\mathcal{K}\|_\infty W_1(PD^{(1)},PD^{(2)}) + 2m\sup \|\nabla \mathcal{K} \|_2 \sum_{l=1}^M\left(\|\mathbf{P}_i^{(1)} - \mathbf{P}_i^{(2)}\|_\infty + \|\mathbf{P}_l^{(1)} - \mathbf{P}_l^{(2)}\|_\infty\right)\\
&=(2\|\mathcal{K}\|_\infty+2m\sup \|\nabla \mathcal{K} \|_2)W_1(PD^{(1)},PD^{(2)})+2mM\sup \|\nabla \mathcal{K} \|_2\|\mathbf{P}_i^{(1)} - \mathbf{P}_i^{(2)}\|_\infty.
\end{aligned}
\end{equation}
}
Finally, we estimate the difference of $\mathbf{Z}^{(1)}$ and $\mathbf{Z}^{(2)}$ under 1-norm as follows:
\begin{equation}
\label{eq:lem4_3_2:7}
\begin{aligned}
\|\mathbf{Z}^{(1)}-\mathbf{Z}^{(2)}\|_1 &= \sum_{i=1}^M \left|z_i^{(1)} - z_i^{(2)}\right|\\
&\leq 2M(m \sup \|\nabla \mathcal{K} \|_2 + \|\mathcal{K}\|_\infty) W_1(PD^{(1)},PD^{(2)}) + 2mM \sum_{i=1}^M\|\mathbf{P}_i^{(1)} - \mathbf{P}_i^{(2)}\|_\infty\\
&=2M(2m\sup \|\nabla \mathcal{K} \|_2+\|\mathcal{K}\|_\infty)W_1(PD^{(1)},PD^{(2)}).
\end{aligned}
\end{equation}
\end{proof}

\begin{proof}
\textbf{of Lemma \ref{lem:matrix_form}.} According to the LSPIA iteration
    method in matrix form in~\pref{eq:iterformat},
    the difference between $\widetilde{\mathbf{Z}}^{(1)}_{(k+1)}$ and $\widetilde{\mathbf{Z}}^{(2)}_{(k+1)}$ under 2-norm is estimated as follows:
{\small
\begin{equation}
\label{eq:lem_matrix_form_1}
\begin{aligned}
&\left\|\widetilde{\mathbf{Z}}^{(1)}_{(k+1)} - \widetilde{\mathbf{Z}}^{(2)}_{(k+1)}\right\|_2\\
& = \left\|\left(\mathbf{E}-\mathbf{\Lambda} (\mathbf{B}^{(1)})^{\rm T} \mathbf{B}^{(1)}\right)\widetilde{\mathbf{Z}}^{(1)}_{(k)} + \mathbf{\Lambda} (\mathbf{B}^{(1)})^{\rm T} \mathbf{Z}^{(1)} - \left(\mathbf{E}-\mathbf{\Lambda} (\mathbf{B}^{(2)})^{\rm T} \mathbf{B}^{(2)}\right)\widetilde{\mathbf{Z}}^{(2)}_{(k)} + \mathbf{\Lambda} (\mathbf{B}^{(2)})^{\rm T} \mathbf{Z}^{(2)}\right\|_2\\
& \leq \left\|\left(\mathbf{E}-\mathbf{\Lambda} (\mathbf{B}^{(1)})^{\rm T} \mathbf{B}^{(1)}\right)\widetilde{\mathbf{Z}}^{(1)}_{(k)} - \left(\mathbf{E}-\mathbf{\Lambda} (\mathbf{B}^{(2)})^{\rm T} \mathbf{B}^{(2)}\right)\widetilde{\mathbf{Z}}^{(2)}_{(k)} \right\|_2 + \left\|\mathbf{\Lambda} (\mathbf{B}^{(1)})^{\rm T} \mathbf{Z}^{(1)} - \mathbf{\Lambda} (\mathbf{B}^{(2)})^{\rm T} \mathbf{Z}^{(2)}\right\|_2,
\end{aligned}
\end{equation}
}
where the second inequity includes two items induced from Property (2) in Lemma \ref{lem_norm}.

First, we consider the first item in \pref{eq:lem_matrix_form_1}.
By adding and subtracting the item
$$\left(\mathbf{E}-\mathbf{\Lambda} (\mathbf{B}^{(1)})^{\rm T} \mathbf{B}^{(1)}\right)\widetilde{\mathbf{Z}}^{(2)}_{(k)}$$
in the first item of \pref{eq:lem_matrix_form_1},
    it follows by Property (2) and (3) in Lemma~\ref{lem_norm} that
{\small
\begin{equation}
\label{eq:lem_matrix_form_2}
\begin{aligned}
&\left\|\left(\mathbf{E}-\mathbf{\Lambda} (\mathbf{B}^{(1)})^{\rm T} \mathbf{B}^{(1)}\right)\widetilde{\mathbf{Z}}^{(1)}_{(k)} - \left(\mathbf{E}-\mathbf{\Lambda} (\mathbf{B}^{(2)})^{\rm T} \mathbf{B}^{(2)}\right)\widetilde{\mathbf{Z}}^{(2)}_{(k)} \right\|_2\\
&= \left\|\left(\mathbf{E}-\mathbf{\Lambda} (\mathbf{B}^{(1)})^{\rm T} \mathbf{B}^{(1)}\right)\left(\widetilde{\mathbf{Z}}^{(1)}_{(k)}-\widetilde{\mathbf{Z}}^{(2)}_{(k)}\right) + \mathbf{\Lambda}\left((\mathbf{B}^{(2)})^{\rm T} \mathbf{B}^{(2)} - (\mathbf{B}^{(1)})^{\rm T} \mathbf{B}^{(1)}\right)\widetilde{\mathbf{Z}}_{(k)}^{(2)}\right\|_2\\
& \leq \left\|\left(\mathbf{E}-\mathbf{\Lambda} ( \mathbf{B}^{(1)})^{\rm T} \mathbf{B}^{(1)}\right)\left(\widetilde{\mathbf{Z}}^{(1)}_{(k)}-\widetilde{\mathbf{Z}}^{(2)}_{(k)}\right) \right\|_2  + \left\|\mathbf{\Lambda}\left( (\mathbf{B}^{(2)})^{\rm T} \mathbf{B}^{(2)} - (\mathbf{B}^{(1)})^{\rm T} \mathbf{B}^{(1)}\right)\widetilde{\mathbf{Z}}_{(k)}^{(2)}\right\|_2\\
& \leq \left\|\mathbf{E}-\mathbf{\Lambda} (\mathbf{B}^{(1)})^{\rm T} \mathbf{B}^{(1)}\right\|_2 \left\|\widetilde{\mathbf{Z}}^{(1)}_{(k)}-\widetilde{\mathbf{Z}}^{(2)}_{(k)}\right\|_2  + \left\|\mathbf{\Lambda}\left( (\mathbf{B}^{(2)})^{\rm T} \mathbf{B}^{(2)} - (\mathbf{B}^{(1)})^{\rm T} \mathbf{B}^{(1)}\right)\right\|_2 \left\|\widetilde{\mathbf{Z}}_{(k)}^{(2)}\right\|_2,
\end{aligned}
\end{equation}
}
According to Property~(4) in Lemma~\ref{lem_norm} and Lemma~\ref{lemma_spect},
    we have
    $$\left\|\mathbf{E}-\mathbf{\Lambda} (\mathbf{B}^{(1)})^{\rm T} \mathbf{B}^{(1)}\right\|_2 = \rho(\mathbf{E}-\mathbf{\Lambda} (\mathbf{B}^{(1)})^{\rm T} \mathbf{B}^{(1)}),$$
    denoted as $\rho_1$.
Additionally, because $\mathbf{\Lambda} = diag(1/C, 1/C, \cdots, 1/C)$, by Property~(1) and (6) in Lemma~\ref{lem_norm},
it holds that
\begin{equation}
\label{eq:lem_matrix_form_3}
\begin{aligned}
& \left\|\mathbf{\Lambda}\left( (\mathbf{B}^{(2)})^{\rm T} \mathbf{B}^{(2)} - (\mathbf{B}^{(1)})^{\rm T} \mathbf{B}^{(1)} \right)\right\|_2 = \left\|\frac{1}{C}\left( (\mathbf{B}^{(2)})^{\rm T} \mathbf{B}^{(2)} - (\mathbf{B}^{(1)})^{\rm T} \mathbf{B}^{(1)}\right)\right\|_2\\
& = \frac{1}{C}\left\|(\mathbf{B}^{(2)})^{\rm T} \mathbf{B}^{(2)} - (\mathbf{B}^{(1)})^{\rm T} \mathbf{B}^{(1)}\right\|_2 \leq \frac{1}{C}\left\| (\mathbf{B}^{(2)})^{\rm T} \mathbf{B}^{(2)} - (\mathbf{B}^{(1)})^{\rm T} \mathbf{B}^{(1)}\right\|_F.
\end{aligned}
\end{equation}
Furthermore, for each entry $b_{ij}^{(\beta)}$ of the matrix ${\mathbf{B}^{(\beta)}}^{\rm T} \mathbf{B}^{(\beta)}$,
    it can be viewed as a function \textit{w.r.t.} the variable
    $\mathbf{u}^{(\beta)} = (s_1^{(\beta)}, s_2^{(\beta)}, \cdots, s_M^{(\beta)},t_1^{(\beta)}, t_2^{(\beta)}, \cdots, t_M^{(\beta)}),\ \beta=1,2$,
    \textit{i.e.},
    $$b_{ij}^{(\beta)} =  b_{ij}(\mathbf{u}^{(\beta)}) = \sum_{l=1}^M B_i(s_l^{(\beta)},t_l^{(\beta)})B_j(s_l^{(\beta)},t_l^{(\beta)}), \ \beta=1,2.$$
Therefore, according to Lemma~\ref{lem4.1}, we have
\begin{equation}
\label{eq:lem_matrix_form_4}
\left|b_{ij}^{(1)} - b_{ij}^{(2)}\right| = \left|b_{ij}(\mathbf{u}^{(1)}) - b_{ij}(\mathbf{u}^{(2)})\right|\leq \sup_{\mathbf{w}}\| \nabla b_{ij}(\mathbf{w})\|_2 \left\|\mathbf{u}^{(1)} - \mathbf{u}^{(2)}\right\|_2,
\end{equation}
where $\mathbf{w} = (s_1, s_2, \cdots, s_M, t_1, t_2, \cdots, t_M) \in [0,1]^{2M}$.
By computing the partial derivative of $b_{ij}(\mathbf{w})$ \textit{w.r.t.} $s_l$,
    we have
\begin{equation}
\label{eq:lem_matrix_form_5}
\frac{\partial b_{ij}(\mathbf{w})}{\partial s_l} = B_i^\prime(s_l)B_i(t_l)B_j(s_l)B_j(t_l) + B_j^\prime(s_l)B_j(t_l)B_i(s_l)B_i(t_l).
\end{equation}
By replacing $s_l, t_l$ with $t_l, s_l$, the partial derivative \textit{w.r.t.} $t_l$ is obtained.
Because cubic B-spline basic function is $C^1$ continuous, the first derivative is bounded in the domain.
Because Lemma~\ref{lem_derivative} gives a bound of the first-order derivative of cubic B-spline basis functions determined by the uniform knot sequence~\pref{eq5},
$\sup_s |B_i^\prime(s)| = \sup_t |B_i^\prime(t)| \leq \alpha$, where $\alpha$ is the bound given in Lemma~\ref{lem_derivative}.
Since $|B_i(s)| \leq 1,\ \text{and}\ |B_i(t)| \leq 1$, by using \pref{eq:lem_matrix_form_5},
    it satisfies that
\begin{equation}
\label{eq:lem_matrix_form_6}
\sup_{\mathbf{w}} \left|\frac{\partial b_{ij}(\mathbf{w})}{\partial s_l}\right| \leq  \sup_{\mathbf{w}}|B_i^\prime(s_l)B_i(t_l)B_j(s_l)B_j(t_l)| + \sup_{\mathbf{w}}|B_j^\prime(s_l)B_j(t_l)B_i(s_l)B_i(t_l)| \leq 2 \alpha.
\end{equation}
Therefore, according to \pref{eq:lem_matrix_form_6}, $\sup_{\mathbf{w}}\|\nabla b_{ij} (\mathbf{w})\|_2 \leq \sqrt{2M \cdot (2\alpha)^2} = 2\sqrt{2M}\alpha$ holds.
It follows by \pref{eq:lem_matrix_form_3} and \pref{eq:lem_matrix_form_4} that
\begin{equation}
\label{eq:lem_matrix_form_7}
\begin{aligned}
& \left\|\mathbf{\Lambda}\left( (\mathbf{B}^{(2)})^{\rm T} \mathbf{B}^{(2)} - (\mathbf{B}^{(1)})^{\rm T} \mathbf{B}^{(1)}\right)\right\|_2 \leq \frac{1}{C}\left\|(\mathbf{B}^{(2)})^{\rm T} \mathbf{B}^{(2)} - (\mathbf{B}^{(1)})^{\rm T} \mathbf{B}^{(1)}\right\|_F\\
& \leq \frac{1}{C} \left( \sum_{i = 1}^{h^2} \sum_{j=1}^{h^2} 8M\alpha^2 \left\| \mathbf{u}^{(1)} - \mathbf{u}^{(2)} \right\|_2^2 \right)^{1/2} = \frac{2\sqrt{2M}\alpha h^2}{C} \left\| \mathbf{u}^{(1)} - \mathbf{u}^{(2)} \right\|_2.
\end{aligned}
\end{equation}
where the size of the matrix $\mathbf{B}^{\rm T} \mathbf{B}$ is $h^2 \times h^2$.

For the item $\left\|\widetilde{\mathbf{Z}}_{(k)}^{(2)}\right\|_2$ in the last equation of \pref{eq:lem_matrix_form_2},
By using \pref{eq:iterformat} and Property (2) and (3) in Lemma~\ref{lem_norm},
    it holds that
\begin{equation}
\label{eq:lem_matrix_form_8}
\begin{aligned}
\left\|\widetilde{\mathbf{Z}}_{(k)}^{(2)}\right\|_2 &=\left\|(\mathbf{E}-\mathbf{\Lambda} (\mathbf{B}^{(2)})^{\rm T} \mathbf{B}^{(2)}) \widetilde{\mathbf{Z}}_{(k-1)}^{(2)} + \mathbf{\Lambda} (\mathbf{B}^{(2)})^{\rm T} \mathbf{Z}^{(2)}\right\|_2\\
& \leq \left\|\mathbf{E}-\mathbf{\Lambda} (\mathbf{B}^{(2)})^{\rm T} \mathbf{B}^{(2)}\right\|_2 \left\|\widetilde{\mathbf{Z}}_{(k-1)}^{(2)}\right\|_2 + \left\|\mathbf{\Lambda} (\mathbf{B}^{(2)})^{\rm T} \right\|_2 \left\|\mathbf{Z}^{(2)}\right\|_2.
\end{aligned}
\end{equation}
According to Property~(4) in Lemma~\ref{lem_norm} and Lemma~\ref{lemma_spect},
    it holds that
    $$\left\|\mathbf{E}-\mathbf{\Lambda} (\mathbf{B}^{(2)})^{\rm T} \mathbf{B}^{(2)}\right\|_2 = \rho(\mathbf{E}-\mathbf{\Lambda} (\mathbf{B}^{(2)})^{\rm T} \mathbf{B}^{(2)}),$$
 denoted as $\rho_2$.
Meanwhile, by using Property~(1) and (5) in Lemma~\ref{lem_norm},
we have
$$ \left\|\mathbf{\Lambda} (\mathbf{B}^{(2)})^{\rm T} \right\|_2 = 1/C \left\|(\mathbf{B}^{(2)})^{\rm T}\right\|_2,$$
and
\begin{equation}
\label{eq:lem_matrix_form_9}
\frac{1}{C} \left\|(\mathbf{B}^{(2)})^{\rm T}\right\|_2^2 = \frac{1}{C}\left\|(\mathbf{B}^{(2)})^{\rm T} \mathbf{B}^{(2)}\right\|_2 = \left\|\mathbf{\Lambda}(\mathbf{B}^{(2)})^{\rm T} \mathbf{B}^{(2)}\right\|_2.
\end{equation}
Because Corollary \ref{col:convergence} shows $\rho(\mathbf{\Lambda}(\mathbf{B}^{(2)})^{\rm T}\mathbf{B}^{(2)}) \leq 1$,
it follows by Property (4) in Lemma \ref{lem_norm} and \pref{eq:lem_matrix_form_9} that $\left\|\mathbf{\Lambda}(\mathbf{B}^{(2)})^{\rm T} \mathbf{B}^{(2)}\right\|_2 \leq 1$,
and $\|(\mathbf{B}^{(2)})^{\rm T}\|_2\leq \sqrt{C}$.
Hence, $ \left\|\mathbf{\Lambda} (\mathbf{B}^{(2)})^{\rm T} \right\|_2 \leq 1/\sqrt{C}$.
By iteratively using \pref{eq:iterformat},
    it is induced from \pref{eq:lem_matrix_form_8} that
\begin{equation}
\label{eq:lem_matrix_form_10}
\begin{aligned}
\left\|\widetilde{\mathbf{Z}}_{(k)}^{(2)}\right\|_2 & \leq \left\|\mathbf{E}-\mathbf{\Lambda} (\mathbf{B}^{(2)})^{\rm T} \mathbf{B}^{(2)}\right\|_2 \left\|\widetilde{\mathbf{Z}}_{(k-1)}^{(2)}\right\|_2 + \left\|\mathbf{\Lambda} (\mathbf{B}^{(2)})^{\rm T} \right\|_2 \left\|\mathbf{Z}^{(2)} \right\|_2\\
&\leq \rho_2 \left\|\widetilde{\mathbf{Z}}_{(k-1)}^{(2)}\right\|_2 + \frac{1}{\sqrt{C}} \left\|\mathbf{Z}^{(2)} \right\|_2\\
&\leq \rho_2^2 \left\|\widetilde{\mathbf{Z}}_{(k-2)}^{(2)} \right\|_2 + \frac{1}{\sqrt{C}}(1+\rho_2) \left\|\mathbf{Z}^{(2)}\right\|_2\ ({\rm using\ \pref{eq:iterformat}\ and\ \pref{eq:lem_matrix_form_8}\ when\ }k=k-1).\\
&\leq \cdots\\
& \leq \rho_2^k \left\|\widetilde{\mathbf{Z}}_{(0)}^{(2)}\right\|_2 + \frac{1}{\sqrt{C}}\left(\sum_{i=0}^{k-1}\rho_2^i\right) \left\|\mathbf{Z}^{(2)}\right\|_2.
\end{aligned}
\end{equation}
Because the initial values of LSPIA are zero, \textit{i.e.}, $\widetilde{\mathbf{Z}}_{(0)}^{(2)} = \mathbf{O}$, and $\rho_2 \leq 1$,
we have $\sum_{i=0}^{k-1}\rho_2^i \leq k$.
By Lemma \ref{lem4.2} that $\|\mathbf{x}\|_2 \leq \sqrt{n} \|\mathbf{x}\|_\infty$ for any $\mathbf{x} \in \mathbb{R}^n$,
$\|\mathbf{Z}^{(2)}\|_2 \leq \sqrt{M} \|\mathbf{Z}^{(2)}\|_\infty$ holds.
It is induced from \pref{eq:lem_matrix_form_10} that
\begin{equation}
\label{eq:lem_matrix_form_11}
\left\|\mathbf{\widetilde{Z}}_{(k)}^{(2)}\right\|_2 \leq k \sqrt{\frac{M}{C}} \left\|\mathbf{Z}^{(2)}\right\|_\infty.
\end{equation}
By \pref{eq:lem_matrix_form_2}, \pref{eq:lem_matrix_form_7}, and \pref{eq:lem_matrix_form_11},
    the first item of \pref{eq:lem_matrix_form_1} can be estimated as follows:
\begin{equation}
\label{eq:lem_matrix_form_12}
\begin{aligned}
&\left\|\left(\mathbf{E}-\mathbf{\Lambda} (\mathbf{B}^{(1)})^{\rm T} \mathbf{B}^{(1)}\right) \widetilde{\mathbf{Z}}^{(1)}_{(k)} - \left(\mathbf{E}-\mathbf{\Lambda} (\mathbf{B}^{(2)})^{\rm T} \mathbf{B}^{(2)}\right) \widetilde{\mathbf{Z}}^{(2)}_{(k)} \right\|_2\\
\qquad \quad& \leq \rho_1 \left\|\widetilde{\mathbf{Z}}^{(1)}_{(k)}-\widetilde{\mathbf{Z}}^{(2)}_{(k)}\right\|_2 + 2\sqrt{2}\alpha h^2 k \frac{M}{\sqrt{C^3}}\left\|\mathbf{Z}^{(2)}\right\|_\infty \left\|\mathbf{u}^{(1)} - \mathbf{u}^{(2)}\right\|_2.
\end{aligned}
\end{equation}

We then consider the second item in \pref{eq:lem_matrix_form_1} by adding and subtracting the same item $\mathbf{\Lambda} (\mathbf{B}^{(1)})^{\rm T} \mathbf{Z}^{(2)}$,
and it then follows by Property~(2) and (3) in Lemma~\ref{lem_norm} that
\begin{equation}
\label{eq:lem_matrix_form_13}
\begin{aligned}
&\left\|\mathbf{\Lambda} (\mathbf{B}^{(1)})^{\rm T} \mathbf{Z}^{(1)} - \mathbf{\Lambda} (\mathbf{B}^{(2)})^{\rm T} \mathbf{Z}^{(2)}\right\|_2 \\
&= \left\|\mathbf{\Lambda} \left((\mathbf{B}^{(1)})^{\rm T} - (\mathbf{B}^{(2)})^{\rm T}\right)  \mathbf{Z}^{(2)} +  \mathbf{\Lambda} (\mathbf{B}^{(1)})^{\rm T} \left( \mathbf{Z}^{(1)} - \mathbf{Z}^{(2)} \right)\right\|_2\\
\qquad & \leq \left\|\mathbf{\Lambda} \left( (\mathbf{B}^{(1)})^{\rm T} - (\mathbf{B}^{(2)})^{\rm T}\right)  \mathbf{Z}^{(2)}\right\|_2 + \left\| \mathbf{\Lambda} (\mathbf{B}^{(1)})^{\rm T} \left( \mathbf{Z}^{(1)} - \mathbf{Z}^{(2)} \right)\right\|_2\\
\qquad & \leq \left\|\mathbf{\Lambda} \left( (\mathbf{B}^{(1)})^{\rm T} - (\mathbf{B}^{(2)})^{\rm T}\right)\right\|_2 \left\| \mathbf{Z}^{(2)}\right\|_2 + \left\| \mathbf{\Lambda} (\mathbf{B}^{(1)})^{\rm T}\right\|_2 \left\|\mathbf{Z}^{(1)} - \mathbf{Z}^{(2)}\right\|_2.
\end{aligned}
\end{equation}
Analogously, by using the technique in \pref{eq:lem_matrix_form_3}, we have
\begin{equation}
\label{eq:lem_matrix_form_14}
\begin{aligned}
\left\|\mathbf{\Lambda} \left( (\mathbf{B}^{(1)})^{\rm T} - (\mathbf{B}^{(2)})^{\rm T}\right)\right\|_2 & \leq \frac{1}{C} \left\| (\mathbf{B}^{(1)})^{\rm T} - (\mathbf{B}^{(2)})^{\rm T} \right\|_F\\
& = \frac{1}{C}\left( \sum_{i=1}^{h^2} \sum_{j=1}^{M} \left|B_i(s_j^{(1)},t_j^{(1)}) - B_i(s_j^{(2)},t_j^{(2)})\right|^2 \right)^{1/2}.
\end{aligned}
\end{equation}
By using Lemma~\ref{lem4.1} on \pref{eq:lem_matrix_form_14},
it holds that
\begin{equation}
\label{eq:lem_matrix_form_15}
\left|B_i(s_j^{(1)},t_j^{(1)}) - B_i(s_j^{(2)},t_j^{(2)})\right| \leq \sup_{(s,t)} \|\nabla B_i(s,t)\|_2 \left\|(s_j^{(1)} - s_j^{(2)}, t_j^{(1)} - t_j^{(2)})\right\|_2,
\end{equation}
where $\sup_{(s,t)} \|\nabla B_i(s,t)\|_2 = \sup_{(s,t)} \left[(B_i^\prime(s)B_i(t))^2 + (B_i(s)B_i^\prime(t))^2\right]^{1/2} \leq \sqrt{2} \alpha$.
Hence, it follows by \pref{eq:lem_matrix_form_14} and \pref{eq:lem_matrix_form_15} that
\begin{equation}
\label{eq:lem_matrix_form_16}
\begin{aligned}
& \|\mathbf{\Lambda} \left( (\mathbf{B}^{(1)})^{\rm T} - (\mathbf{B}^{(2)})^{\rm T}\right)\|_2\\
& \leq \frac{1}{C} \left[ \sum_{i=1}^{h^2} \sum_{j=1}^{M}2\alpha^2 \left((s_j^{(1)} - s_j^{(2)})^2 + (t_j^{(1)} - t_j^{(2)})^2 \right) \right]^{1/2}\\
& \leq \frac{1}{C} \left[  \sum_{j=1}^{M}2\alpha^2 h^2 \left((s_j^{(1)} - s_j^{(2)})^2 + (t_j^{(1)} - t_j^{(2)})^2\right) \right]^{1/2}\\
& \leq \frac{\sqrt{2}\alpha h}{C} \left[  \sum_{j=1}^{M} \left((s_j^{(1)} - s_j^{(2)})^2 + (t_j^{(1)} - t_j^{(2)})^2\right) \right]^{1/2}\\
& \leq \frac{\sqrt{2}\alpha h}{C} \left\|\mathbf{u}^{(1)} - \mathbf{u}^{(2)}\right\|_2.
\end{aligned}
\end{equation}
Moreover, by the same technique used in \pref{eq:lem_matrix_form_9},
it can be shown that $ \left\|\mathbf{\Lambda} (\mathbf{B}^{(1)})^{\rm T} \right\|_2 \leq 1/\sqrt{C}$.
Since $\|\mathbf{Z}^{(2)}\|_2 \leq \sqrt{M}\|\mathbf{Z}^{(2)}\|_\infty$,
it follows by \pref{eq:lem_matrix_form_13} and \pref{eq:lem_matrix_form_16} that
\begin{equation}
\label{eq:lem_matrix_form_17}
\left\|\mathbf{\Lambda} (\mathbf{B}^{(1)})^{\rm T} \mathbf{Z}^{(1)} - \mathbf{\Lambda} (\mathbf{B}^{(2)})^{\rm T} \mathbf{Z}^{(2)}\right\|_2 \leq \sqrt{2}\alpha h\frac{\sqrt{M}}{C} \|\mathbf{Z}^{(2)}\|_\infty \left\|\mathbf{u}^{(1)} - \mathbf{u}^{(2)}\right\|_2 + \frac{1}{\sqrt{C}} \left\|\mathbf{Z}^{(1)} - \mathbf{Z}^{(2)}\right\|_2.
\end{equation}

Finally, it follows by \pref{eq:lem_matrix_form_1}, \pref{eq:lem_matrix_form_12}, and \pref{eq:lem_matrix_form_17} that
{\small
\begin{equation}
\label{eq:lem_matrix_form_18}
\left\|\widetilde{\mathbf{Z}}^{(1)}_{(k+1)} - \widetilde{\mathbf{Z}}^{(2)}_{(k+1)}\right\|_2 \leq \rho_1 \left\|\widetilde{\mathbf{Z}}^{(1)}_{(k)}-\widetilde{\mathbf{Z}}^{(2)}_{(k)}\right\|_2 + \sqrt{2}\alpha h(2h k \frac{M}{\sqrt{C^3}}+\frac{\sqrt{M}}{C})\|\mathbf{Z}^{(2)}\|_\infty \left\|\mathbf{u}^{(1)} - \mathbf{u}^{(2)}\right\|_2 + \frac{ \left\|\mathbf{Z}^{(1)} - \mathbf{Z}^{(2)} \right\|_2}{\sqrt{C}}.
\end{equation}
}
By iteratively using \pref{eq:iterformat} and \pref{eq:lem_matrix_form_18} when the iteration is $k-1, k-2 ,\cdots, 1$,
while keeping the item $2hk$ in the inequity because $2h(k-i) \leq 2hk$ for $i = 1,2, \cdots, k-1$,
we have
{\small
\begin{equation}
\label{eq:lem_matrix_form_19}
\begin{aligned}
&\left\|\widetilde{\mathbf{Z}}^{(1)}_{(k+1)} - \widetilde{\mathbf{Z}}^{(2)}_{(k+1)}\right\|_2 \\
&\leq \rho_1^2 \left\|\widetilde{\mathbf{Z}}^{(1)}_{(k-1)}-\widetilde{\mathbf{Z}}^{(2)}_{(k-1)}\right\|_2 + (1+\rho_1) \left(\sqrt{2}\alpha h(2h k \frac{M}{\sqrt{C^3}}+\frac{\sqrt{M}}{C})\|\mathbf{Z}^{(2)}\|_\infty \left\|\mathbf{u}^{(1)} - \mathbf{u}^{(2)}\right\|_2 + \frac{\left\|\mathbf{Z}^{(1)} - \mathbf{Z}^{(2)}\right\|_2}{\sqrt{C}} \right)\\
& \leq \cdots\\
& \leq \rho_1^{k+1} \left\|\widetilde{\mathbf{Z}}^{(1)}_{(0)}-\widetilde{\mathbf{Z}}^{(2)}_{(0)}\right\|_2 + \left( \sum_{i=0}^{k} \rho_1^i \right) \left(\sqrt{2}\alpha h(2h k \frac{M}{\sqrt{C^3}}+\frac{\sqrt{M}}{C})\|\mathbf{Z}^{(2)}\|_\infty \left\|\mathbf{u}^{(1)} - \mathbf{u}^{(2)}\right\|_2 + \frac{\left\|\mathbf{Z}^{(1)} - \mathbf{Z}^{(2)}\right\|_2}{\sqrt{C}} \right).
\end{aligned}
\end{equation}
}
Due to Lemma~\ref{lemma_spect}, $\rho_1 \leq 1$.
Thus, $ \sum_{i=0}^{k} \rho_1^i \leq k+1$ holds.
$\|\mathbf{Z}^{(2)}\|_\infty$ can be controlled by the maximum of the eminence function $\mathcal{E}$, \textit{i.e.}, $\|\mathbf{Z}^{(2)}\|_\infty \leq \|\mathcal{E}\|_\infty$.
Meanwhile, because the initial values of LSPIA are all zero,
$\widetilde{\mathbf{Z}}^{(i)}_{(0)} = \mathbf{O},\ i=1,2$.
Hence, the stability result is induced from \pref{eq:lem_matrix_form_19} that
\begin{equation*}
\left\|\widetilde{\mathbf{Z}}^{(1)}_{(k+1)} - \widetilde{\mathbf{Z}}^{(2)}_{(k+1)}\right\|_2
\leq \sqrt{2}\alpha h(k+1)(2h k \frac{M}{\sqrt{C^3}}+\frac{\sqrt{M}}{C})\|\mathcal{E}\|_\infty \left\|\mathbf{u}^{(1)} - \mathbf{u}^{(2)}\right\|_2 + \frac{k+1}{\sqrt{C}} \left\|\mathbf{Z}^{(1)} - \mathbf{Z}^{(2)}\right\|_2,
\end{equation*}
where the iteration is $k+1$, and $\mathcal{E}$ represents the eminence function.
\end{proof}

\section{Appendix: Convergence and Computational Complexity of LSPIA}
\label{Appendix_B}
In our theoretical framework, the constant $C$ is larger than $ M^2$, where $M$ is the number of points in a PD.
In practical implementation, to accelerate the convergence of LSPIA, $C$ is set to be $\|\mathbf{B}^{\rm T}\|_\infty$.
To clarify the convergence of LSPIA, the following theorem is summarized from  \cite{lin2013efficient}, \cite{deng2014progressive} and \cite{lin2018convergence}.

\begin{theorem}
\label{thm:convergence1}
 Let $\mathbf{B}$ be the B-spline collocation matrix given
     by~\pref{eq:matrixB},
    and $\mathbf{B}^\mathrm{T}\mathbf{B}\widetilde{\mathbf{Z}} = \mathbf{B}^\mathrm{T}\mathbf{Z}$ be the normal equation~\pref{eq:normal_equation} of the least-square fitting problem~\pref{eq:min_leastsquare}.
If the spectral radius $\rho(\mathbf{\Lambda B^{\rm T} B})\leq 1$,
    then whatever $\mathbf{B}^\mathrm{T}\mathbf{B}$ is singular or not, the LSPIA is convergent:
    \begin{itemize}
     \item[(1)] If $\mathbf{B}^\mathrm{T}\mathbf{B}$ is non-singular, the iterative solution given in \pref{eq:iterformat} converges to the solution of the linear system, i.e., $(\mathbf{B}^\mathrm{T}\mathbf{B})^{-1}\mathbf{B}^{\mathrm{T}}\mathbf{Z}$;
     \item[(2)] If $\mathbf{B}^\mathrm{T}\mathbf{B}$ is singular,
    the iterative method shown in \pref{eq:iterformat} converges to the M-P pseudo-inverse solution of the linear system. Moreover, if the initial value $\mathbf{\widetilde{Z}}_{(0)} = \mathbf{O}$, the iterative method converges to
    $(\mathbf{B}^\mathrm{T}\mathbf{B})^{+}\mathbf{B}^{\mathrm{T}}\mathbf{Z}$,
    \textit{i.e.}, the M-P pseudo-inverse solution of the linear system with the minimum Euclidean norm,
    where $(\mathbf{B}^\mathrm{T}\mathbf{B})^{+}$ represents the M-P pseudo-inverse of $\mathbf{B}^\mathrm{T}\mathbf{B}$.
    \end{itemize}
\end{theorem}
    Now, we show that the LSPIA format we used to generate PBSGs converges. Specifically, it needs to verify the condition $\rho(\mathbf{\Lambda} \mathbf{B}^\mathrm{T}\mathbf{B})\leq1$.
Hence, we have the following corollary according to the theorem above.

\begin{corollary}
\label{col:convergence}
The LSPIA format~\pref{eq:iterformat} to generate PBSGs converges.
\end{corollary}
\begin{proof}
The iterative format is rewritten in matrix form with the initial condition $\widetilde{\mathbf{Z}}_{(0)}$:
    $$\widetilde{\mathbf{Z}}_{(p+1)} =(\mathbf{E} - \mathbf{\Lambda} \mathbf{B}^\mathrm{T} \mathbf{B}) \widetilde{\mathbf{Z}}_{(p)} + \mathbf{\Lambda} \mathbf{B}^{\mathrm{T}} \mathbf{Z},$$
    where $\mathbf{\Lambda} = diag(1/C, 1/C\cdots, 1/C)$ is a diagonal matrix, and $C\geq \|\mathbf{B}^{\rm T}\|_\infty$ in the format.
According to Theorem~\ref{thm:convergence1},
    it only needs to verify $\rho(\mathbf{\Lambda} \mathbf{B}^\mathrm{T}\mathbf{B})\leq1$.

We consider $\|\mathbf{\Lambda} \mathbf{B}^{\mathrm{T}} \mathbf{B}\|_\infty$.
First,
    because the B-spline basis has the weight property, i.e.,
    $B_i(s_l)B_j(t_l) \leq 1$, for $i,j = 1,2,\cdots h$,
	$||\mathbf{B}||_\infty = 1$ holds,
	where $|| \cdot ||_\infty$ denotes the induced $\infty$-norm of a matrix.
Second, since
\begin{equation*}
\begin{aligned}
\sum_{l = 1}^{M}\frac{B_i(s_l) B_j(t_l)}{C} &\leq \sum_{l = 1}^{M}\frac{B_i(s_l) B_j(t_l)}{M^2} \leq \sum_{l = 1}^{M}\frac{B_i(s_l) B_j(t_l)}{M} \leq \frac{\sum_{l = 1}^{M}B_i(s_l) B_j(t_l)}{\|\mathbf{B}^{\rm T}\|_\infty}\\
& = \frac{\sum_{l = 1}^{M}B_i(s_l) B_j(t_l)}{\max_{1\leq i,j \leq h}\left\{ \sum_{l = 1}^{M} B_i(s_l) B_j(t_l)\right\}} \leq 1, M\geq 1,
\end{aligned}
\end{equation*}
    we have $|| \mathbf{\Lambda} \mathbf{B}^\mathrm{T} ||_\infty \leq 1$.
Therefore,
    $$\rho(\mathbf{\Lambda} \mathbf{B}^\mathrm{T} \mathbf{B} ) \leq \| \mathbf{\Lambda} \mathbf{B}^\mathrm{T} \mathbf{B} \|_\infty \leq \| \mathbf{\Lambda} \mathbf{B}^\mathrm{T} \|_\infty ||\mathbf{B}||_\infty \leq 1.$$
According to Theorem~\ref{thm:convergence1},
	the LSPIA format to generate PBSGs is convergent.
\end{proof}

Additionally, a lemma is provided to show $\rho(\mathbf{E} - \mathbf{\Lambda}\mathbf{B}^{\rm T}\mathbf{B})\leq 1$,
    which is used to prove the stability theorems in Section~\ref{stab}.
\begin{lemma}
\label{lemma_spect}
  Let $\mathbf{B}$ be the B-spline collocation matrix given by~\pref{eq:matrixB} and
  $$\mathbf{\Lambda} = diag(1/C, 1/C\cdots, 1/C)$$
  is a diagonal matrix, with $C\geq \|\mathbf{B}^{\rm T}\|_\infty$.
  The eigenvalues of the matrix $\mathbf{\Lambda}\mathbf{B}^{\rm T}\mathbf{B}$ are all real and not less than zero, and then $\rho(\mathbf{E} - \mathbf{\Lambda}\mathbf{B}^{\rm T}\mathbf{B})\leq 1$.
\end{lemma}

\begin{proof}
Suppose that $\lambda$ is an arbitrary eigenvalue of the matrix $\mathbf{\Lambda}\mathbf{B}^{\rm T}\mathbf{B}$ with eigenvector $\mathbf{x}$, \textit{i.e.},
\begin{equation}
\label{eq:lemma_spect_1}
    \mathbf{\Lambda}\mathbf{B}^{\rm T}\mathbf{B}\mathbf{x} = \lambda \mathbf{x}.
\end{equation}
Then, by multiplying $\mathbf{B}$ at both sides of \pref{eq:lemma_spect_1},
    it holds that
    $$\mathbf{B}\mathbf{\Lambda}\mathbf{B}^{\rm T}(\mathbf{B}\mathbf{x}) = \lambda (\mathbf{B}\mathbf{x}). $$
It shows that $\lambda$ is also an eigenvalue of the matrix $\mathbf{B}\mathbf{\Lambda}\mathbf{B}^{\rm T}$ with eigenvector $\mathbf{B}\mathbf{x}$.
Since for any $\mathbf{y} \in \mathbb{R}^M$,
    $$\mathbf{y}^{\rm T}\mathbf{B}\mathbf{\Lambda}\mathbf{B}^{\rm T}\mathbf{y} = \mathbf{y}^{\rm T}\mathbf{B}\mathbf{\Lambda}^{1/2} (\mathbf{\Lambda}^{1/2})^{\rm T}\mathbf{B}^{\rm T}\mathbf{y} = (\mathbf{y}^{\rm T}\mathbf{B}\mathbf{\Lambda}^{1/2})(\mathbf{y}^{\rm T}\mathbf{B}\mathbf{\Lambda}^{1/2})^{\rm T}\geq 0, $$
    the matrix $\mathbf{B}\mathbf{\Lambda}\mathbf{B}^{\rm T}$ is a positive semidefinite matrix. Therefore, $\lambda$ is real and $\lambda\geq 0$.

Moreover, according to Corollary~\ref{col:convergence}, $\rho(\mathbf{\Lambda} \mathbf{B}^\mathrm{T} \mathbf{B})\leq 1$,
    the eigenvalues $\lambda$ of $\mathbf{\Lambda} \mathbf{B}^\mathrm{T} \mathbf{B}$ satisfy $0\leq \lambda \leq 1$.
Furthermore, the eigenvalues $\lambda^\prime$ of the matrix $\mathbf{E} - \mathbf{\Lambda} \mathbf{B}^\mathrm{T} \mathbf{B}$ satisfy $\lambda^\prime = 1-\lambda$,
    and thus we have $0\leq \lambda^\prime \leq 1$, and $\rho(\mathbf{E} - \mathbf{\Lambda}\mathbf{B}^{\rm T}\mathbf{B})\leq 1$.
\end{proof}

Finally, the computational complexity and memory consumption per iteration step of the LSPIA method are provided.
Each iteration step of the LSPIA method requires $(2M+1)h^2$ multiplications and $2Mh^2$ additions.
Meanwhile, it is required to store two $h^2\times 1$ vectors, $\widetilde{\mathbf{Z}}_{(k)}$ and $\widetilde{\mathbf{Z}}_{(k+1)}$,
    an $M\times 1$ vector $\mathbf{Z}$, and an $M\times h^2$ matrix $\mathbf{B}$. Therefore, each iteration requires $(M+2)h^2 + M$ unit memory.

    \vskip 0.2in
    \bibliography{PB_ref}

\end{document}